\documentclass[final,3p,times,numbers,compress]{elsarticle}

\graphicspath{{figure/}}
\usepackage{amsmath}
\usepackage{subfigure}
\usepackage{graphicx}
\usepackage{mathrsfs}
\usepackage{caption}
\usepackage{ragged2e}
\usepackage{float}
\usepackage{subcaption}
\usepackage{array}
\usepackage{setspace}
\usepackage{threeparttable}

\usepackage {CJK}
\usepackage[utf8]{inputenc}
\usepackage[T1]{fontenc}
\usepackage{mathptmx}
\usepackage{color,xcolor}
\usepackage{booktabs}
\biboptions{compress}

\journal{Sustainable Cities and Society}
\begin{document}

\include{UDSymbol}

\captionsetup[figure]{labelfont={bf},labelformat={default},labelsep=period,name={Fig.}}

\begin{frontmatter}

\title{Generalization of Urban Wind Environment Using Fourier Neural Operator Across Different Wind Directions and Cities}

%
\author[label1]{Cheng Chen}
\author[label1]{Geng Tian}
\author[label3]{Shaoxiang Qin}
\author[label1]{Dingyang Geng}
\author[label1]{Dongxue Zhan}
\author[label1]{Jinqiu Yang}
\author[label3]{David Vidal}
\author[label1]{Liangzhu Wang\corref{cor1}}

\address[label1]{Concordia University, Department of Building, Civil and Environmental Engineering, and Department of Computer Science and Software Engineering, Montreal, Canada}
\address[label2]{McGill University, School of Computer Science, Montreal, Canada}
\address[label3]{Polytechnique Montreal, Department of Mechanical Engineering, Montreal, Canada}

\cortext[cor1]{Email address for correspondence: leon.wang@concordia.ca}

\begin{abstract}

Simulation of urban wind environments is crucial for urban planning, pollution control, and renewable energy utilization. However, the computational requirements of high-fidelity computational fluid dynamics (CFD) methods make them impractical for real cities. To address these limitations, this study investigates the effectiveness of the Fourier Neural Operator (FNO) model in predicting flow fields under different wind directions and urban layouts. In this study, we investigate the effectiveness of the Fourier Neural Operator (FNO) model in predicting urban wind conditions under different wind directions and urban layouts. By training the model on velocity data from large eddy simulation data, we evaluate the performance of the model under different urban configurations and wind conditions. The results show that the FNO model can provide accurate predictions while significantly reducing the computational time by 99\%. Our innovative approach of dividing the wind field into smaller spatial blocks for training improves the ability of the FNO model to capture wind frequency features effectively. The SDF data also provides important spatial building information, enhancing the model’s ability to recognize physical boundaries and generate more realistic predictions. The proposed FNO approach enhances the AI model’s generalizability for different wind directions and urban layouts.
\end{abstract}


\end{frontmatter}


\section{Introduction}
\label{sec:level1}

In recent years, with the rapid growth of population and accelerated urbanization, natural landscapes have gradually been transformed into dense urban environments dominated by various building structures. This transformation has led to the formation of urban microclimates, which are localized atmospheric conditions that differ significantly from those of the surrounding natural areas. These microclimates are characterized by changes in wind speed and direction, temperature, and altered pollutant dispersion patterns, all of which are deeply influenced by the complex interaction between urban structures and atmospheric flow. Understanding and predicting these dynamic phenomena is crucial for effective urban planning, public health, and environmental management \cite{amirtham2009mapping,smargiassi2009variation,oke2006towards,yang2023urban}. The primary methods for studying urban microclimates include field observations \cite{bulkeley2023condition,aminipouri2019urban,wang2021cool}, wind tunnel experiments \cite{reynolds_measurements_2008, blocken2016pedestrian, blackman_assessment_2019}, and numerical simulations \cite{wang2021recent,liu2017determination,hager2014parallel,tian_impact_2024}.  
The field observation method, conducted in natural environments, captures realistic climatic conditions and accounts for complex factors such as solar radiation, humidity, and wind speed variations \cite{santamouris2001impact}. However, it requires significant resources, limiting its scope to small-scale data collection. Wind tunnel experiments accurately simulate airflow around buildings but are costly and subject to interference from tunnel walls. Compared to these methods, CFD is efficient and flexible and has become widely used in microclimate simulations \cite{wright2003non,ding2022efficient,shen2023impact,solazzo2008modelling,zheng2021cfd,pesic2016large,mortezazadeh2022cityffd}.  Direct numerical simulation (DNS) demands high computational resources, restricting it to low-Reynolds number research \cite{yeung2001lagrangian,coleman2003direct,awad2022assessment,sarkar1991direct}. 
Large-eddy simulation (LES) approach balances computational requirements with high accuracy, which is more practical for atmospheric flow analysis \cite{li2008large, park_large-eddy_2015, yoshida_large-eddy-simulation_2018,tian_turbulence-kinetic-energy_2021, tian_note_2023}. Due to the high computational cost and slow simulation speed, many researchers in urban wind field simulation have sought new tools for faster predictions or to supplement CFD simulations, partially replacing CFD in fluid mechanics analysis. With advancements in GPU hardware, GPU-based large-eddy simulation (LES) software, such as CityFFD \cite{mortezazadeh2022cityffd}, has demonstrated strong capabilities for large-scale numerical modeling. Additionally, the availability of larger training datasets has driven increasing interest in AI technologies and data-driven neural network models.

This research aims to bridge the gap between traditional computational methods and AI-based approaches in urban wind field prediction. As urbanization accelerates, scalable and efficient models are essential. Traditional methods like CFD are accurate but computationally intensive, limiting real-time application in large urban environments. Integrating deep learning with CFD offers faster predictions without compromising accuracy, supporting efficient urban planning and sustainable management. The ability to generalize predictions across various urban layouts can benefit cities globally, making advanced simulations accessible to diverse industrial and urban development needs.

Raissi et al. \cite{raissi2019physics} introduced Physics-Informed Neural Networks (PINNs) for solving Partial Differential Equations(PDEs), integrating physics-based loss terms into neural network training. While PINNs excel in inverse problems \cite{cai2021physics} and flow field completion \cite{li2024investigation, chen_three-dimensional_2025}, they offer limited speed advantages over traditional CFD methods for forward problems and show weaker generalization across PDE families \cite{cuomo2022scientific}. A key benefit of PINNs is their ability to train without real simulation training data by directly embedding PDE residuals into the loss function. However, their slower performance in forward tasks has driven interest toward alternative AI methods that provide faster data processing and improved generalization. Data-driven models \cite{rudy2017data} have emerged as promising options, trained on PDE initial conditions and solutions to uncover underlying physical relationships.
In light of these considerations, the neural operator method has emerged as a focal point in the research community.  The neural operator method \cite{kovachki2023neural} addresses the PDEs between data by learning solutions in infinite-dimensional function spaces. The main neural operators currently available are divided into three categories based on how the function simulation is implemented:

1) \textbf{Deep Operator Network}: The DeepONet model \cite{lu2019deeponet} proposed by Lu et al. approximates operations by splitting the input into two branches: one branch learns the representation of the function, and the other branch captures the specific points of the function evaluation. 
PI-DeepONet \cite{diab2024learning} proposed by Diab et al. represents a pioneering contribution to the field of deep operators. 
It incorporates physical errors during training and employs unsupervised learning without relying on training data, thereby demonstrating that deep neural operators are also capable of unsupervised learning.

2) \textbf{Approximation of a function in Physical space}: 
This approach employs the flexibility of neural networks in physical space (e.g., Graph Neural Network(GNN)\cite{jiang2022graph}, Convolutional Neural Network(CNN)\cite{kattenborn2021review}) to model intricate relationships in data. Convolutional Neural Operator(CNO), proposed by Raonic et al.\cite{raonic2023convolutional} combines U-Net and CNN for the first time and performs feature extraction based on the pure spatial domain, which is superior to FNO in terms of supporting data diversity. Pangu, proposed by Bi et al. \cite{bi2022pangu}, outperforms numerical weather prediction (NWP) models, is trained using 4D data for the first time, and integrates multiple physical information.  However, Pangu requires a high training time cost and a huge amount of training data, which limits its usage in wide applications.

3) \textbf{Approximation of a function in Fourier space}: this approach uses the Fourier transform to model an integral operator in spectral space to capture global dependencies.
Li et al.\cite{kovachki2024operator} proposed the FNO method based on the Fourier transform, which improves the computational efficiency while significantly improving the performance on multiple PDE tasks compared to previous mainstream PDE AI methods. FourCastNet proposed by Pathak et al.\cite{pathak2022fourcastnet} is based on the AFNO framework. AFNO proposed by Guibas et al.\cite{guibas2021adaptive} is based on FNO and employs a block-wise structure to adaptively share weights across channels in channel mixing weights. Furthermore, it specifies the frequency mode through the use of soft thresholding and contraction. AFNO demonstrates superior performance to self-attention mechanisms in terms of both efficiency and accuracy for few-shot segmentation tasks. FourCastNet is capable of providing accurate short- to medium-term global high-resolution and rapidly changing multi-meteorological variable predictions with a resolution of 0.25\textdegree. While the FNO model can generate a one-week forecast in under two seconds, it requires a large amount of training data. Therefore, we decide to adopt FNO in our research.
In summary, this study addresses the following key aspects of research on micro-urban wind field simulation:

1. \textbf{High prediction accuracy and efficiency}: Using the FNO model, the 2D wind field at different wind directions at the same height in the same city can achieve prediction accuracy close to that of mainstream CFD software, and the prediction time of a single iteration is less, thus it can assist the CFD method in simulating micro-urban wind fields.

2. \textbf{Memory-efficient training approach}: Due to the high GPU memory demands when training over an entire urban area with full modes and large width, we opted for a small-piece training strategy. This approach divides the urban wind field into smaller segments, allowing training to proceed effectively even with constrained GPU resources. Unlike whole-piece training, which struggles to capture global frequency characteristics due to memory constraints, the small-piece training method can fully capture the frequency features of the input wind field, leading to better overall performance under the same hardware configuration.

3. \textbf{Generalization across urban geometries}: The trained FNO model demonstrates strong predictive performance in urban wind fields with geometries similar to those of the training data. Specifically, the FNO model requires training on a city environment with a particular geometric configuration and can then achieve high-accuracy predictions in other urban layouts with similar geometric structures. This enhances the FNO model's applicability and scalability across different but geometrically similar urban scenarios.

In the following sections, the research problem will be defined in section \ref{PF}. Next, in section \ref{methodology}, the research methodology will be described, including the use of the FNO model and CityFFD. Section \ref{Numeri} will focus on the generation of training data, detailing the processes used to obtain and preprocess the datasets for the experiments. In section \ref{exp}, the results will be presented and analyzed, comparing the performance of the proposed model with traditional methods. Section \ref{discuss} will discuss the limitations of this study and explore potential directions for future work. Finally, Section \ref{conclu} concludes the paper by summarizing the key findings and contributions of the investigation.

\section{Problem formulation}
\label{PF}

\begin{figure}
    \centering
\includegraphics[scale=0.4]{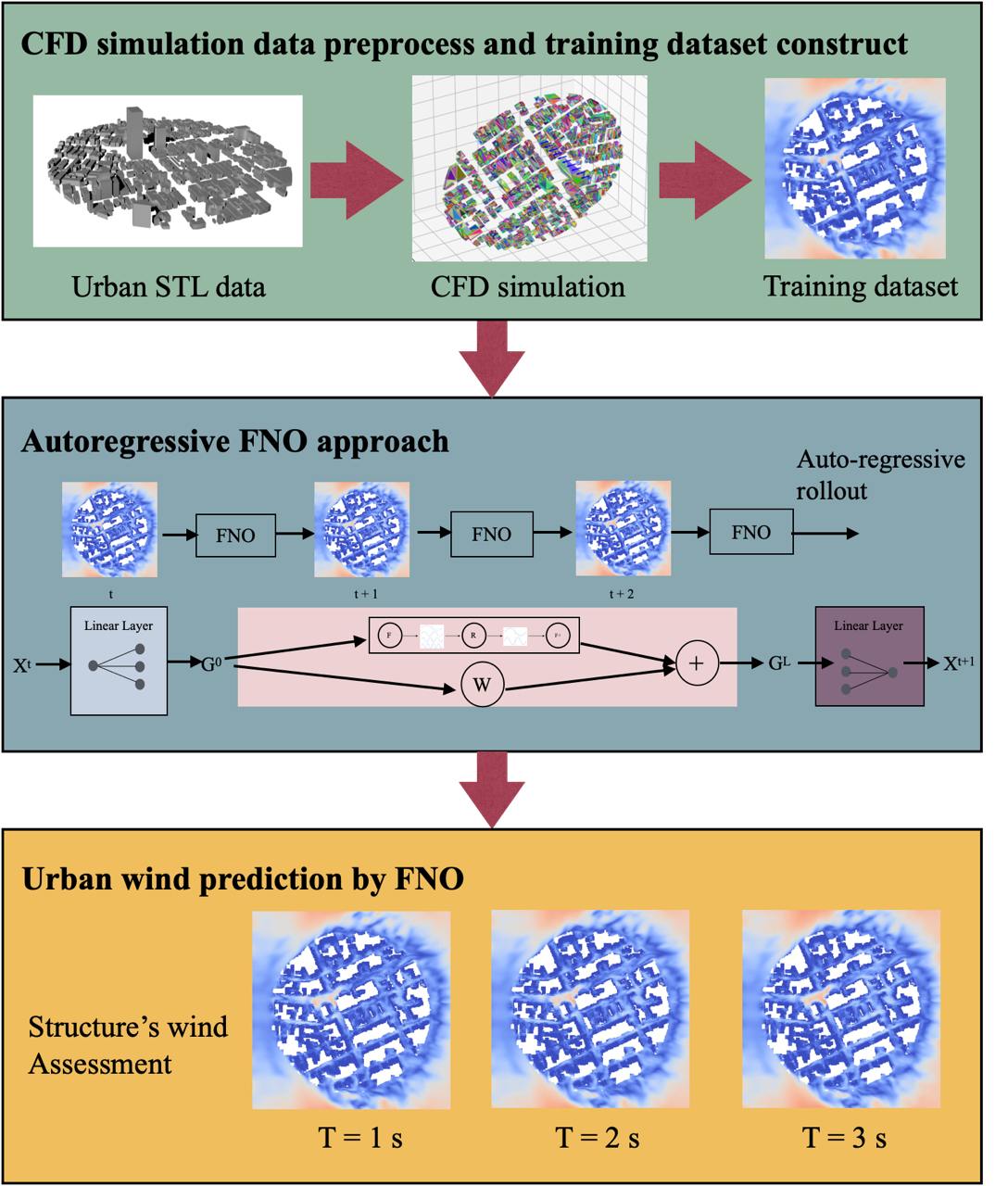}
    \caption{Integration of CityFFD data with the autoregressive FNO approach to wind field prediction within urban-scale buildings.}
    \label{fig_Integration}
\end{figure}

Accurate and efficient urban wind field simulations are essential for applications such as urban planning, environmental monitoring, and wind energy assessment. While traditional Computational Fluid Dynamics (CFD) methods are highly accurate, they are computationally expensive and time-intensive, especially in densely built urban areas with complex geometries. This challenge is magnified when accounting for dynamic changes, such as varying wind directions or evolving urban structures, which require repeated simulations for accurate results. To address these challenges, this study integrates the CityFFD solver with an autoregressive FNO approach to achieve near real-time predictions of wind fields across congested urban landscapes. As shown in Fig.\ref{fig_Integration}, the proposed workflow consists of three main stages: data preprocessing and training dataset generation via CFD simulations, deployment of an autoregressive FNO model, and urban wind field prediction. In the first stage, 3D urban models in STL format serve as input for CFD simulations, generating high-resolution training data that captures the intricate wind patterns around urban structures. In the second stage, the autoregressive FNO model iteratively predicts future wind fields using global features extracted via Fourier transforms combined with a memory-efficient autoregressive rollout. This approach allows the FNO model to learn the temporal dynamics of urban wind patterns while reducing computational demands. In the final stage, the trained FNO model predicts wind flow across multiple future time (e.g., $T=0.5s, 1.0s, 1.5s$), providing near-instantaneous wind predictions for densely built urban areas. The integration of the CityFFD solver with the autoregressive FNO framework addresses the limitations of traditional CFD methods, offering a scalable solution that balances accuracy with computational efficiency. This method has potential for real-time applications in urban design and environmental policy.

\section{Methodology}
\label{methodology}

\subsection{Building data preprocess}

In this study, we used the CityFFD software to perform large-eddy simulations of urban wind fields, generating comprehensive datasets for training and testing the FNO network. CityFFD, equipped with semi-Lagrangian methods and fractional stepping techniques and further accelerated by GPU technology, efficiently simulated microclimatic characteristics in large urban areas. To capture a range of architectural scenarios, we selected Niigata and Montreal as representative urban environments, offering diverse building geometries and layouts. The urban architectural data for these simulations were derived from STL (Standard Triangle Language) format models. STL, commonly used in 3D printing and computer-aided design, provides highly accurate surface geometry representations of three-dimensional objects. In our context, STL data allowed CityFFD to access precise building outlines and structural details, enhancing the accuracy and reliability of the simulations. To seamlessly integrate these intricate 3D models into CFD simulations, we implemented a rigorous data preprocessing protocol, ensuring that every detail of the urban landscape was represented accurately. This careful preparation was crucial to the success of the simulations and, ultimately, to the robustness of the FNO model’s predictions:

\textbf{Step 1:}  Initially, the geographic coordinates in the STL files (latitude and longitude) were converted to a projected coordinate system. This transformation standardized units to meters (m), facilitating compatibility with CFD simulation requirements.

\textbf{Step 2:} Subsequently, the transformed coordinate data were used to construct three-dimensional building models compatible with CFD numerical modeling. This involved precisely converting the contour lines and height data from each building into three-dimensional forms.

\textbf{Step 3:} Based on the spatial layout of the buildings, a uniform grid system was created. This grid provided the necessary spatial framework for CFD simulations and also established the foundational graph structure for the FNO network. This setup ensured high-resolution and detailed simulation fidelity.

Through these preparatory steps, CityFFD could successfully simulate wind flow in Niigata and Montreal, providing high-quality data for training and testing of the FNO network. The accuracy of the CityFFD simulations was verified against traditional CFD tools, demonstrating its ability to simulate urban microclimates in complex environments. These precise simulation results provided a solid foundation for this study.

\subsection{Model configuration}

\begin{figure}[b!]
\centering
\includegraphics[scale=0.25]{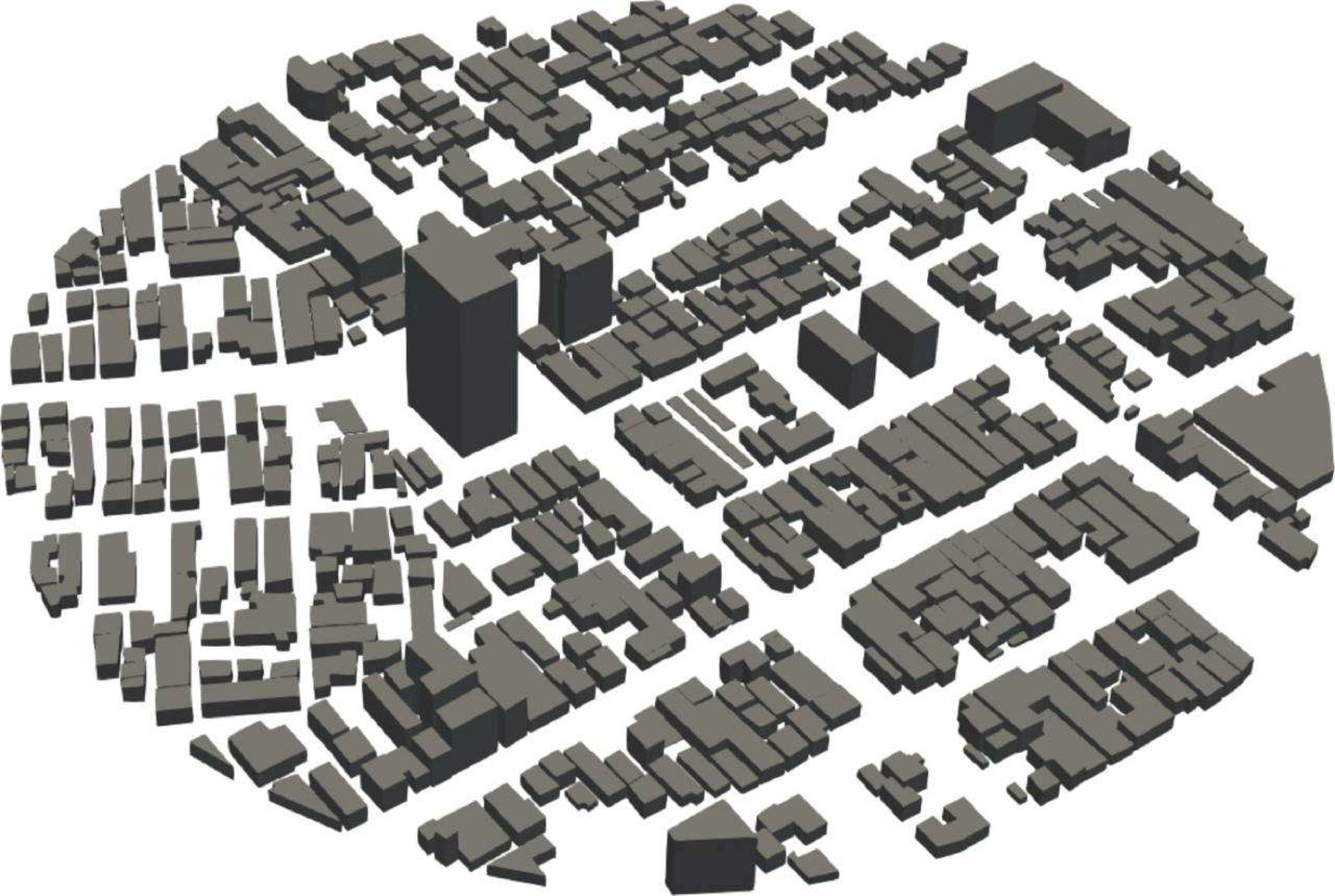}
\caption{Visualization of the building layout in Niigata, Japan.}
\label{Fig.NiigataCaseStudy}
\end{figure}

\begin{table}
\centering
\caption{Summary of the wind field, SDF dataset characteristics, FNO model hyper-parameters, computational time, and parameters scale.}
\resizebox{\textwidth}{!}{  
{\fontsize{5pt}{7pt}\selectfont
\begin{tabular}{@{}l@{\hskip 50pt}l@{}}
\toprule
\textbf{\fontsize{5pt}{7pt}\selectfont Characteristic} & \textbf{\fontsize{5pt}{7pt}\selectfont Details} \\
\midrule
\textbf{\fontsize{5pt}{7pt}\selectfont Wind Field Training Dataset} & \\
\midrule
Spatial Resolution           & $256 \times 256$  / $64 \times 64$ \\
Temporal Resolution (Time Steps) & 1080 \\
Wind Speed Range (m/s)         & Up to 7.8 \\
Wind Direction                 & West \\
Data Coverage                  & 50\% \\
Altitude Levels (m)            & 2 meters \\
Data Format                    & .npy \\
Total Dataset Size (GB)        & 1.8 GB ($64 \times 64$) / 5.6 GB ($256 \times 256$) \\
\midrule
\textbf{\fontsize{5pt}{7pt}\selectfont SDF (Signed Distance Function) Training Dataset} & \\
\midrule
Spatial Resolution             & $256 \times 256$ / $64 \times 64$ \\
Encoding Method                & Distance-based \\
Geometric Representation       & Urban Structures (buildings, roads) \\
Coordinate System              & Cartesian \\
Data Coverage                  & 50\% \\
Data Format                    & .npy \\
Total Dataset Size (GB)        & 770.2 MB ($64 \times 64$) / 251.5 MB ($256 \times 256$) \\
\midrule
\textbf{\fontsize{5pt}{7pt}\selectfont FNO Model Hyper-parameters} & \\
\midrule
Number of Layers               & 4 \\
Activation Function            & GeLU \\
Modes                          & 32 \\
Width                          & 64 \\
Batch Size                     & 100 \\
Input Length (time steps)      & 5 \\
Output Length (time steps)     & 10  \\
\midrule
\textbf{\fontsize{5pt}{7pt}\selectfont Computational Time Comparison} & \\
\midrule
\textbf{\fontsize{5pt}{7pt}\selectfont Method} & \textbf{\fontsize{5pt}{7pt}\selectfont GPU Time (s)} \\
\midrule
CityFFD  & 2.210 \\
FNO      & 0.006 \\
\midrule
\textbf{\fontsize{5pt}{7pt}\selectfont Parameter Comparison} & \\
\midrule
\textbf{\fontsize{5pt}{7pt}\selectfont Method} & \textbf{\fontsize{5pt}{7pt}\selectfont Scale} \\
\midrule
CityFFD  & 24,000,000 mesh elements \\
FNO      & 33,091,306 parameters \\
\bottomrule
\end{tabular}
}}
\label{tab_combined_wind_sdf_hyperparams_performance}
\end{table}

\begin{figure}
\centering
\includegraphics[scale=0.26]{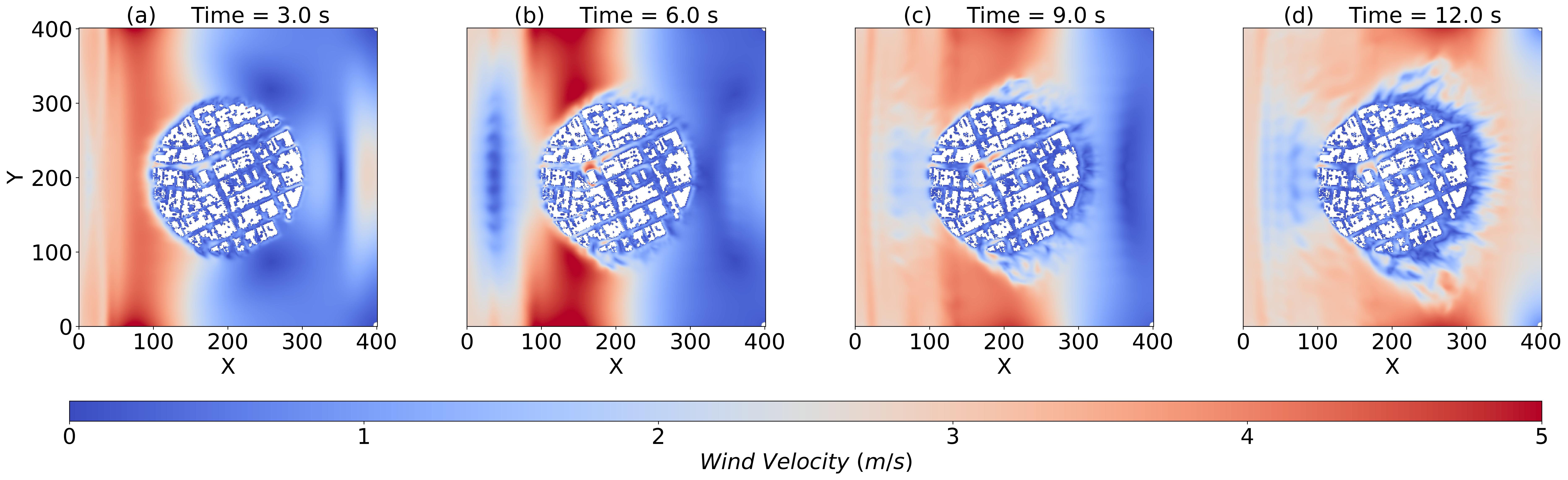}
\caption{Visualization of the wind field evolution generated by CityFFD at time = 3.0 s, 6.0s, 9.0s and 12.0s over Niigata, Japan.}
\label{Fig.evo}
\end{figure}

The study selected wind field simulation data from Niigata city, as shown in Fig.\ref{Fig.NiigataCaseStudy}. To reduce computational complexity while preserving urban characteristics, a simplified building model was adopted to represent the actual city environment. The building area has a radius of 200 meters, covering a computational domain of 400 $\times$ 400 meters, providing a precise geometric basis for the wind field simulation. In the simulation setup, the wind boundary layer height was set to 250 meters, with typical wind directions of west and north to capture the primary wind characteristics in the urban area. A total grid number of 24 million was used to ensure a detailed capture of complex flow structures between buildings. The reference wind speed was set at 7.8 m/s, with an input height of 2 meters for simulation data.
The height of 2 meters was chosen for training data input because it approximates pedestrian height, reflecting wind speeds experienced by individuals in a real urban environment. This height allows researchers to better simulate and evaluate the near-ground wind speed distribution around buildings, particularly concerning pedestrian comfort and safety. By training the FNO model at this height, the study aims to capture urban wind field characteristics more accurately, providing targeted guidance for optimizing pedestrian wind environments and urban planning.
A comprehensive summary of the wind field dataset, SDF dataset, FNO model hyperparameters, and the computational performance comparison between the CityFFD and FNO models is provided in Table \ref{tab_combined_wind_sdf_hyperparams_performance}. The wind field dataset is characterized by a spatial resolution of either $256 \times 256$ or $64 \times 64$, 1080 time steps, a maximum wind speed of 7.8 m/s, a westward wind direction, and a height of 2 meters, stored in .npy format with file sizes of 1.8 GB ($64 \times 64$) and 5.6 GB ($256 \times 256$). The SDF dataset, encoding urban geometry through a distance-based method, matches the spatial resolution of the wind dataset, with file sizes of 770.2 MB ($64 \times 64$) and 251.5 MB ($256 \times 256$). Time steps from t = 0.1s to t = 0.5s were selected as inputs for training, with outputs iterated up to t = 15s. 
Figure \ref{Fig.evo} shows the progression of the wind field above the building layout in Niigata over time. Computational fluid dynamics (CFD) simulations were used to visualize the distribution of wind speeds at the starting time (3.0 seconds) and how it changes every 3 seconds afterward. The initial wind field configuration was chosen as the basis for analysis, focusing on the spin-up phase to understand the flow's transient evolution from the beginning to the fully developed state. The dataset was then divided into 80\% for training and 20\% for validation to facilitate model training and assessment.

Through hyperparameter tuning, the optimal model configuration was determined to be 32 modes, a width of 64, and a 4-layer network with GeLU activation, allowing the FNO model to balance complexity with computational efficiency and leverage the GeLU function for improved gradient stability and convergence. The FNO model was configured to use 5 input time steps to generate 10 output steps, enabling short-term dynamics to be captured and supporting longer-term forecasts. A 50\% data coverage rate was selected for both the wind field and SDF training datasets to prevent overfitting, enhancing the FNO model’s generalization. After 100 epochs, strong generalization to varying wind conditions was demonstrated by the FNO model, which achieved a reduction in computation time to 0.006 seconds per step compared to CityFFD's 2.210 seconds. Despite comprising over 33 million parameters, the FNO model significantly outperformed CityFFD in computational efficiency, making it well-suited for real-time urban wind field predictions.

\subsection{Training method}

\begin{figure}
\centering

\begin{minipage}{0.45\textwidth}
    \centering
    \includegraphics[scale=0.6]{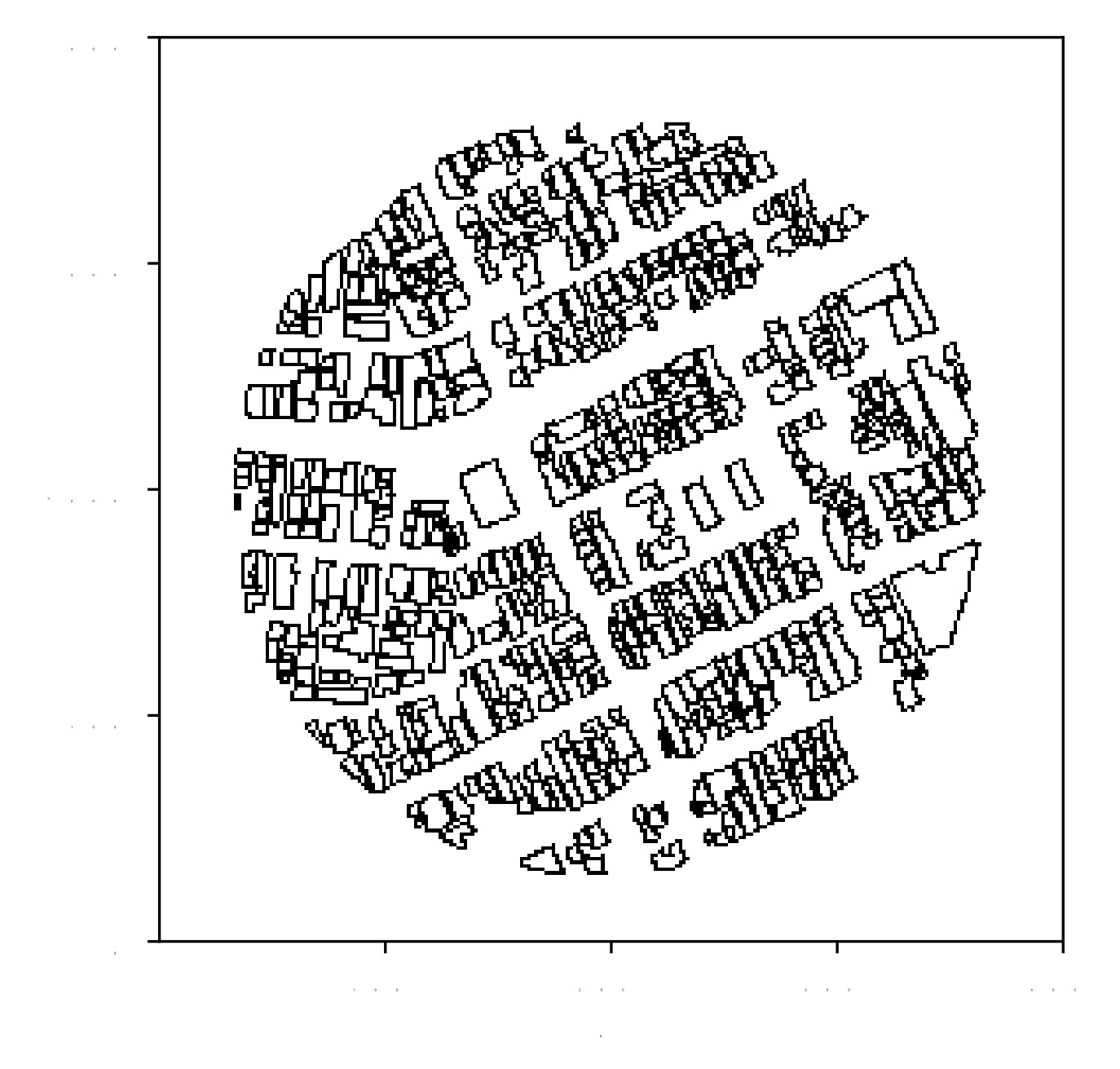}
    \put(-200,200){(a)}
\end{minipage}%
\hspace{0.02\textwidth}
\begin{minipage}{0.45\textwidth}
    \centering
    \includegraphics[scale=0.6]{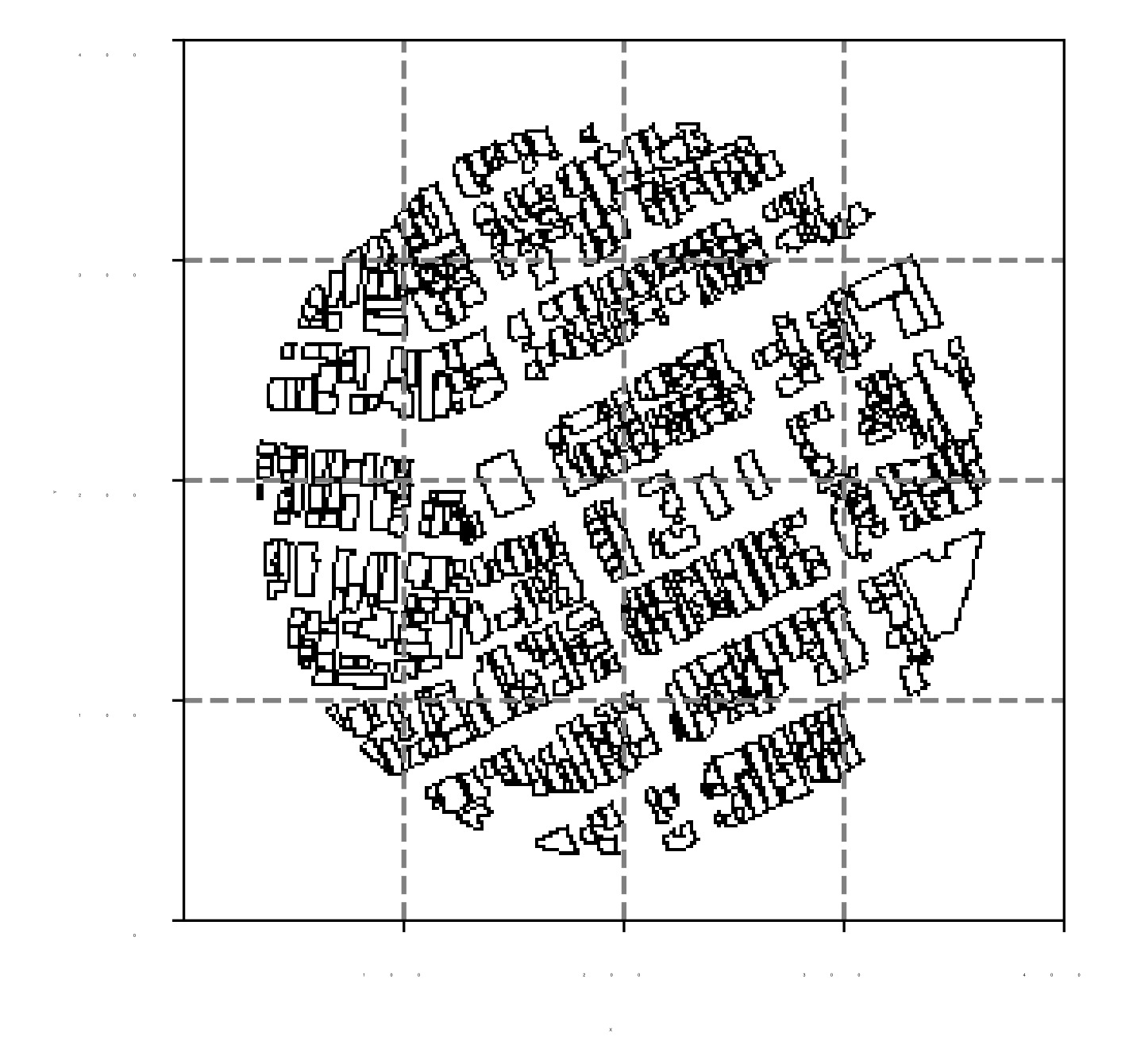}
    \put(-200,200){(b)}
\end{minipage}

\hspace{0.03\textwidth} 
\begin{minipage}{0.45\textwidth}
    \includegraphics[scale=0.44]{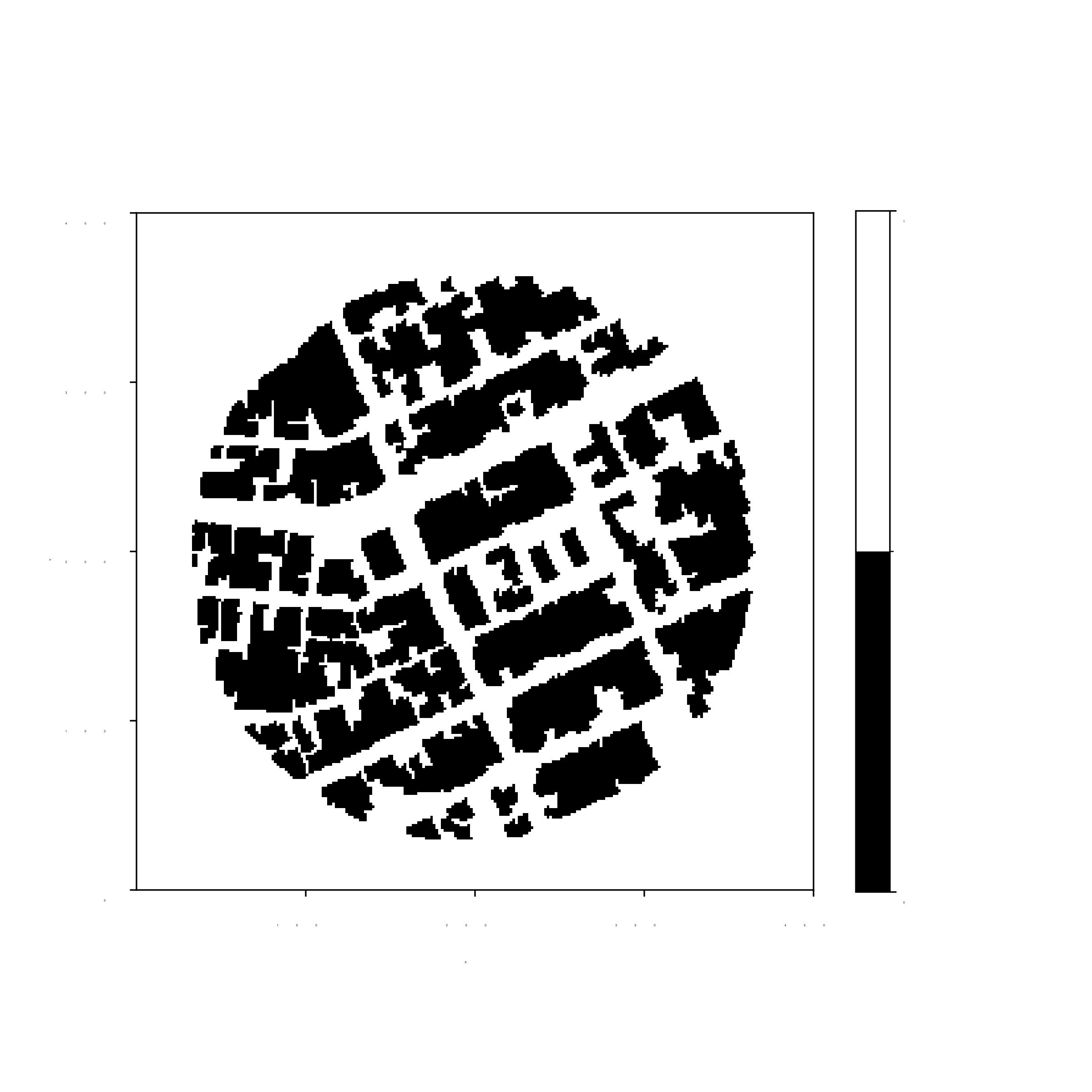}
    \put(-240,223){(c)} 
\end{minipage}%
\hspace{0.03\textwidth} 
\begin{minipage}{0.45\textwidth}
    \centering
    \includegraphics[scale=0.51]{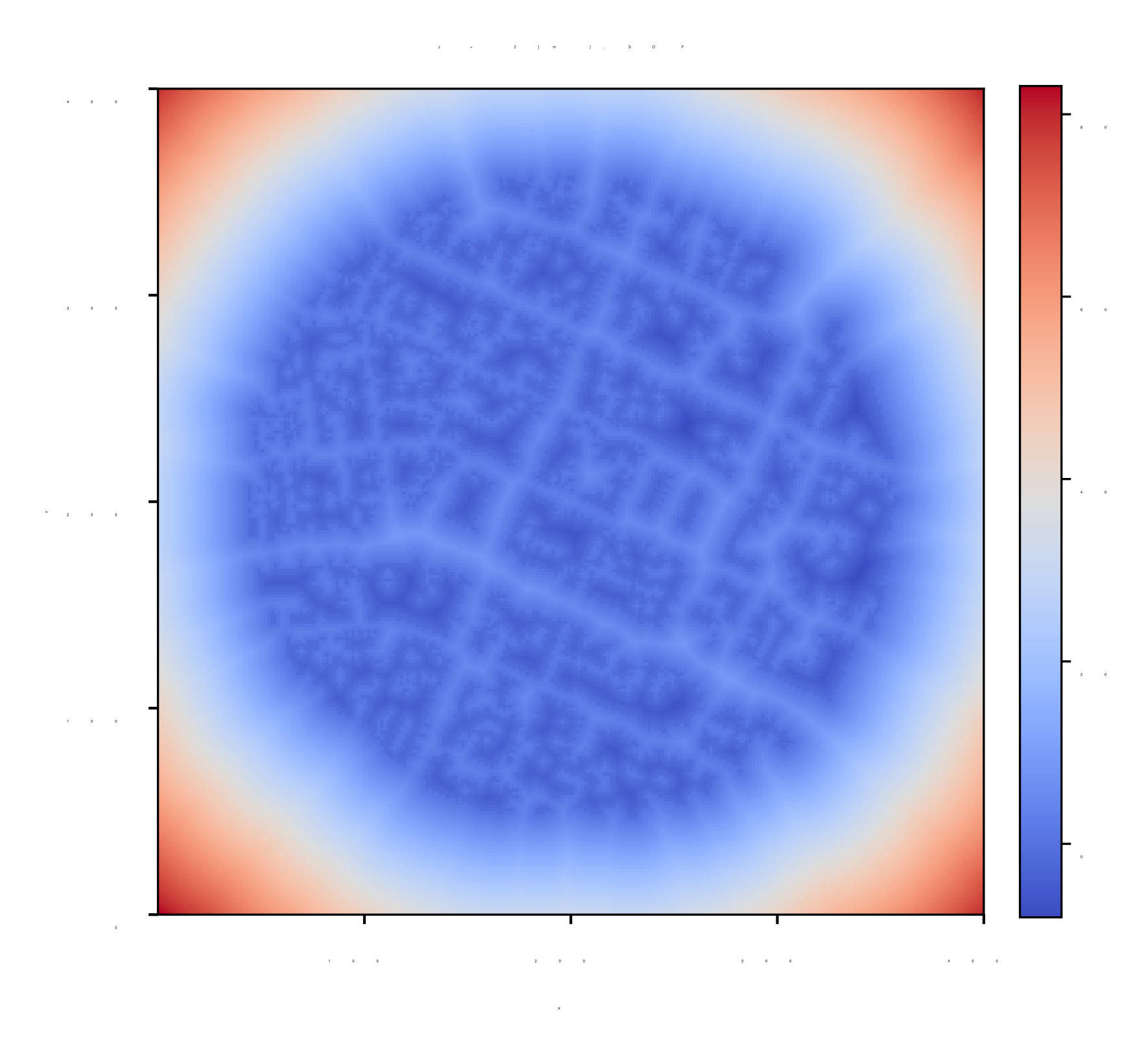}
    \put(-215,200){(d)} 
\end{minipage}

\caption{\textbf{(a)} the original whole urban area layout, \textbf{(b)} the layout segmented into 64 $\times$ 64 patches for FNO-based generalization tasks, \textbf{(c)} Niigata build mask, \textbf{(d)} Niigata SDF.}
\label{Fig.patchesVsWholeDivisionAndMasks}
\end{figure}

To investigate the impact of different training methods on the performance of wind field simulations in urban environments, we present a series of comparative visualizations in Fig.\ref{Fig.patchesVsWholeDivisionAndMasks}. This figure illustrates different training methods for the Niigata urban wind field simulation, providing a visual comparison of the approaches. The top row compares segmentation versus non-segmentation in the training process. Fig.\ref{Fig.patchesVsWholeDivisionAndMasks}~(a) shows the complete urban layout of Niigata without segmentation, highlighting the spatial arrangement of buildings and roads. Fig.\ref{Fig.patchesVsWholeDivisionAndMasks}~(b) depicts the segmented version of the same area, divided into smaller 64 $\times$ 64 patches using a dashed grid. This comparison aims to observe the impact of segmentation on the FNO model’s ability to learn localized wind field features, potentially improving computational efficiency and model accuracy. The second row compares the use of building masks and the derived Signed Distance Function (SDF). Fig.\ref{Fig.patchesVsWholeDivisionAndMasks}~(c) displays a binary building mask, where black regions indicate buildings and white regions represent open spaces. Fig.\ref{Fig.patchesVsWholeDivisionAndMasks}~(d) shows the SDF derived from this mask, with a gradient indicating distances from building boundaries, where darker colors represent areas closer to buildings. This comparison focuses on understanding the role of incorporating SDF data into the FNO model training process, aiming to better capture the influence of urban geometry on wind patterns. These visualizations explore how different preprocessing techniques might influence the FNO model's ability to generalize across various urban environments.

\section{Numerical benchmark datasets}
\label{Numeri}
\subsection{Numerical modeling}
The training and testing datasets are generated from large-eddy simulations using a GPU-based software CityFFD \cite{mortezazadeh2020solving}. CityFFD utilizes a semi-Lagrangian approach and a fractional stepping method, leveraging the power of GPU to simulate urban microclimate characteristics for large-scale urban modeling. The underlying theory of CityFFD has been detailed in previous studies \cite{mortezazadeh2022cityffd,mortezazadeh2019adaptive,mortezazadeh2017high}, and it has been applied in urban contexts to assess wind flow \cite{mortezazadeh2022estimating} and thermal conditions \cite{katal2022urban,katal2019modeling}. Additionally, CityFFD can be coupled with the Weather Research and Forecasting (WRF) model to simulate urban wind distribution \cite{wang2023evaluating,mortezazadeh2021integrating}, and it can also be integrated with urban building energy simulations \cite{luo2022data}.

CityFFD solves the following incompressible Navier–Stokes equations (Eq.\ref{eq:data}), where $\tilde{U}$, $\theta$, and $p$, $Re$, $Gr$, $Pr$, $v_t$, and $\alpha_t$ represent the velocity, temperature, pressure, Reynolds number, Grashof number, Prandtl number, turbulent viscosity, and turbulent thermal diffusivity, respectively:
\begin{equation}
\label{eq:data}
\begin{aligned}
    & \nabla \cdot \tilde{U} = 0 \\
    & \frac{\partial \tilde{U}}{\partial t} + \left ( \tilde{U} \cdot \nabla \right ) \tilde{U} = - \nabla p + \left ( \frac{1}{Re} + \nu_t \right ) \nabla^2 \tilde{U} - \frac{Gr}{Re^2} \theta \\
    & \frac{\partial \theta}{\partial t} + \left ( \tilde{U} \cdot \nabla \right ) \theta = \left ( \frac{1}{Re \cdot Pr} +\alpha_t \right ) \nabla^2 \theta\\
\end{aligned}
\end{equation}

The detection terms in Eq.\ref{eq:data} are addressed using the Lagrangian method, where values for the unknown variables (e.g., velocity and temperature) at position $S_c^{n+1}$ are determined by calculating the values of $\tilde{U}$ and $\theta$ at position $S_c^n$, as illustrated in Eq.\ref{eq:n}.
\begin{equation}
\label{eq:n}
    S_c = \tilde{U} \mathrm{d} t \rightarrow S_c^n \approx S_c^{n+1} - \tilde{U}\Delta t
\end{equation}
and the eddy viscosity is represented by the following equation:
\begin{equation}
\label{eq:turbulence}
v_t = (c_s \Delta)^2 |\tilde{S}|
\end{equation}
The Smagorinsky constant $c_s$ typically ranges from 0.1 to 0.24 and varies based on the type of flow \cite{jiang_natural_2003, tian_new_2020}. Here, $\Delta$ represents the grid size, and $\tilde{S} =\sqrt{2S_{ij}S_{ij}}$ where $S_{ij}$ is the rate-of-strain tensor. More detailed information on the method can be found in previous literature \cite{mortezazadeh2019cityffd}.  CityFFD has been extensively validated against CFD benchmarks, showing satisfactory accuracy for urban microclimate simulations under both isothermal and non-isothermal conditions \cite{mortezazadeh2019cityffd,mortezazadeh2022cityffd}. By considering boundary conditions such as wind speed profiles, wind direction, turbulent viscosity, air temperature, ground temperature, and building surface temperature, CityFFD effectively predicts wind speed and air temperature.

The numerical simulation in this study focuses on a specific area in Niigata, Japan, with a diameter of 400 m. This area has been previously studied and validated in other research works \cite{tominaga2008aij,mortezazadeh2022cityffd}. The computational domain is a square measuring 800 m by 800 m with an elevation range of 300 m. A grid with a spatial resolution of 2 m is utilized, resulting in 24 million grid cells. Wind speeds from the west are used for validation purposes. More detailed validation information on CityFFD can be found in our previous publication \cite{mortezazadeh2022cityffd}. The vertical wind velocity profile follows a power law with a reference velocity of 200 m and a power law exponent of 0.25, consistent with previous studies. The simulation results have been validated for urban airflow simulations in the past.
To collect the necessary training data for the FNO model, we simulated the 3D wind field in the Niigata area using CityFFD. The simulations involved 1,200 continuous-time steps, each representing 0.1 seconds, starting with a westerly wind direction. Convergence criteria were established based on the reference wind speed at the inlet, with wind speed being monitored. We set the threshold for residuals at 0.001 and the correlation coefficient at 0.4. Our focus was on assessing pedestrian wind comfort around buildings, so we gathered wind data at a height of 2 meters above the buildings.

\subsection{Training dataset preprocessing}
This study utilized wind field data for the Niigata region with a resolution of $256 \times 256$, covering 1020 time steps at intervals of 0.1 seconds, resulting in a total simulation duration of 102 seconds. The simulation data, created using CityFFD and saved in \texttt{vtk} format, needed to be converted into a numpy format with dimensions $256 \times 256 \times 1020$ to be compatible with the FNO model. The urban wind field dataset and relevant parameters are summarized in Table \ref{tab_combined_wind_sdf_hyperparams_performance}. To address GPU memory limitations, a sliding window technique was employed to segment the continuous wind velocity data into 15-step windows, with 5 steps as input and 10 steps as output. By sliding the window forward by 2 steps at each iteration, the FNO model could effectively capture localized features and dynamic variations in the wind field while maintaining continuity in the time series.
The FNO model utilized the wind velocity vector magnitude as input for time steps $t$ to $t+4$, calculated by taking the square root of the sum of the squares of the wind velocity components in the $x$ and $y$ directions. The output of the model predicted wind velocity magnitudes for time steps $t+5$ to $t+14$. The FNO model operated in an auto-regressive manner, using predictions from the previous 5 time steps as input for subsequent iterations. Training data consisted of westerly wind data from the Niigata region, while the test set included northerly wind data from Niigata, vertically flipped westerly wind data from Niigata, and westerly wind data from Montreal. To mitigate error accumulation over time steps, Signed Distance Function (SDF) data were integrated into the training process. Hyperparameter optimization was also conducted to enhance training efficiency and improve prediction accuracy within the limitations of available hardware and data.

\begin{table}[H]
\centering
\caption{Summary of cases across different training methods, urban geometries, and wind directions.}

\resizebox{\textwidth}{!}{  
\fontsize{6pt}{8pt}\selectfont

\setlength{\heavyrulewidth}{0.1mm}   
\setlength{\lightrulewidth}{0.05mm}  
\setlength{\aboverulesep}{0.7pt}     
\setlength{\belowrulesep}{0.7pt}     
\setlength{\arrayrulewidth}{0.05mm}   

\resizebox{\textwidth}{!}{
\begin{tabular}{c@{\extracolsep{50pt}}c@{\extracolsep{50pt}}c}

\toprule
\textbf{Comparison Types} & \textbf{Cases}  & \textbf{Instruction} \\
\midrule
\multicolumn{3}{l}{\textbf{Training Methods (Section 5.1)}} \\
\midrule
Train Dataset
                          & \textit{W-Nii-T-CFD} & Train in total domain without SDF\\
                          & \textit{W-Nii-T-SDF-CFD} & Train in total domain with SDF\\
                          & \textit{W-Nii-P-CFD} & Train in patches without SDF\\
                          & \textit{W-Nii-P-SDF-CFD} & Train in patches with SDF\\
\midrule
Test Cases
                          & \textit{W-Nii-T} & Test in total domain without SDF \\
                          & \textit{W-Nii-T-SDF}  & Test in total domain with SDF \\
                          & \textit{W-Nii-P} & Test in patches without SDF\\
                          & \textit{W-Nii-P-SDF}  & Test in patches with SDF\\
\midrule
\multicolumn{3}{l}{\textbf{Wind Directions Generalization (Section 5.2)}} \\
\midrule
Train Dataset
                          & \textit{W-Nii-P-SDF-CFD} & Train in Niigata with west wind\\
\midrule
Test Cases
                          & \textit{N-Nii-P-SDF}   & Test in different wind direction\\
                          & \textit{N-Nii-P-SDF-R} & Rotate 90$^\circ$ counterclockwise\\
\midrule
\multicolumn{3}{l}{\textbf{Urban Geometries Generalization (Section 5.3)}} \\
\midrule
Train Dataset
                          & \textit{W-Nii-P-SDF-CFD} & Train in Niigata with west wind\\
    \midrule
Test Cases
                          & \textit{W-Mon-P-SDF}   & Test in different city geometry\\
                          & \textit{W-Nii-P-SDF-VF} & Vertically flipped geometry\\
\bottomrule
\end{tabular}
}

}

\vspace{10pt}  
\parbox{\textwidth}{\footnotesize
\textbf{*} \textbf{\textit{W}} in the first column stands for West wind direction, \textbf{\textit{N}} for North wind direction, \textbf{\textit{Nii}} for Niigata, \textbf{\textit{Mon}} for Montreal, \textbf{\textit{SDF}} for Signed distance function data, \textbf{\textit{CFD}} for Computational fluid dynamics data, \textbf{\textit{P}} for patch blocks for train, and \textbf{\textit{T}} for total domain for train, \textbf{\textit{R}} for Rotated 90 degrees counterclockwise, \textbf{\textit{VF}} for Vertically flipped. 

}

\label{tab:cases_summary_combined}
\end{table}

\section{Results}
\label{exp}

\subsection{Evaluation of different training methods}

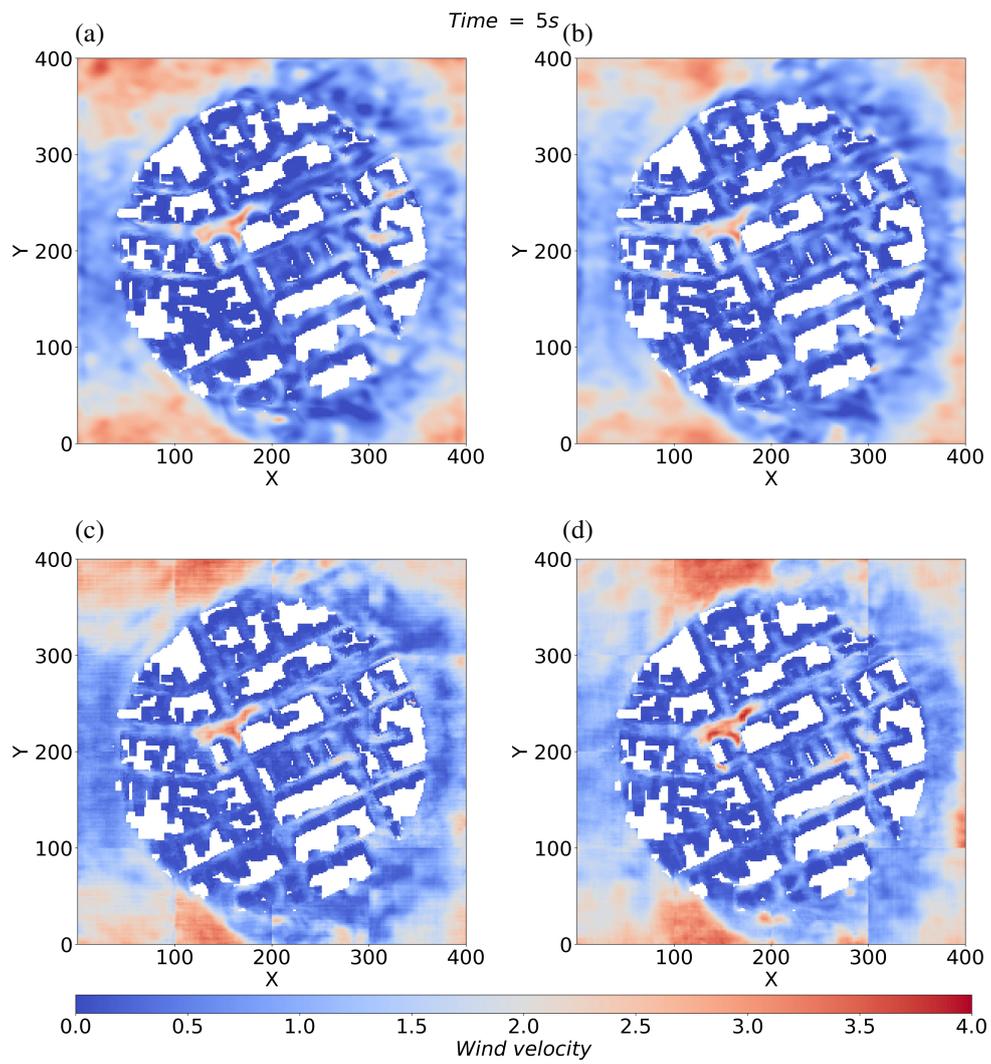
\begin{figure}
\centering

\includegraphics[scale=0.25]{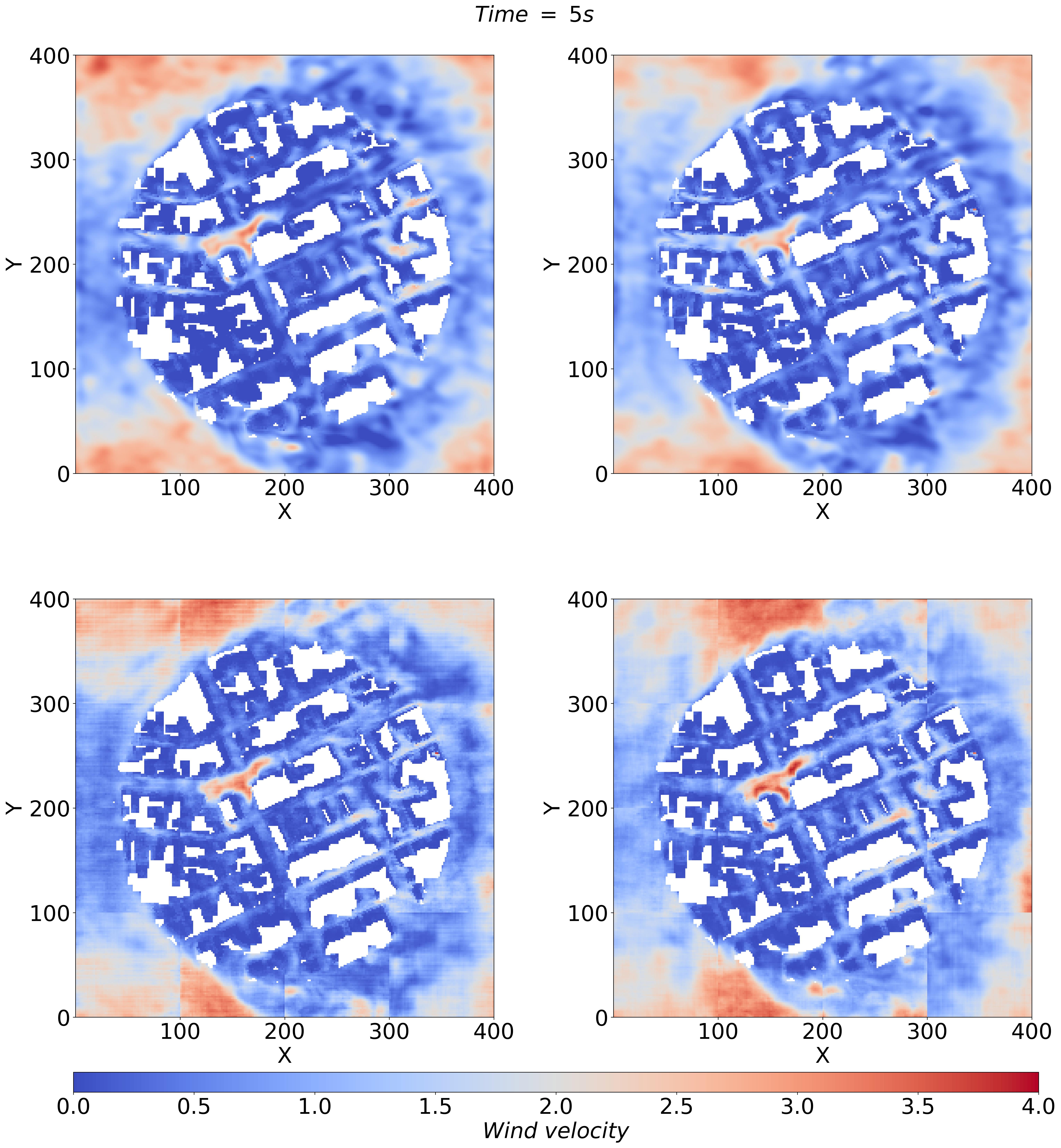}

\setlength{\unitlength}{1\textwidth} 
\begin{picture}(0,0)
    \put(-0.34,0.85){(a)}
    \put(0.05,0.85){(b)}
    \put(-0.34,0.45){(c)}
    \put(0.05,0.45){(d)}
\end{picture}

\caption{Visualization of FNO outputs using different training strategies: (a) \textit{W-Nii-T}, (b) \textit{W-Nii-T-SDF}, (c) \textit{W-Nii-P}, (d) \textit{W-Nii-P-SDF}.}
\label{Fig.T5-1Visual}
\end{figure}

\begin{figure}[H]
\centering
\includegraphics[scale=0.06]{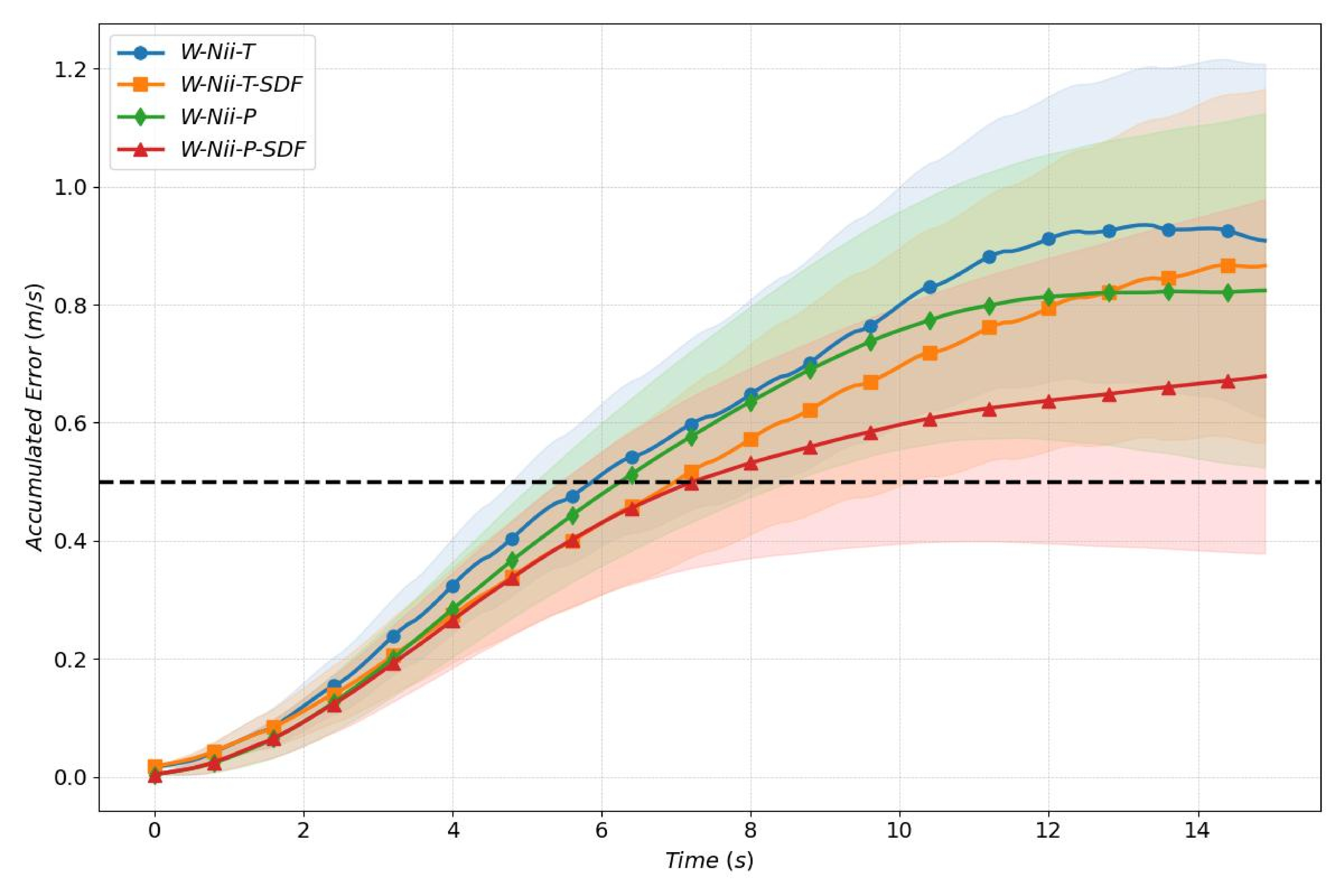}
\caption{Comparison of accumulated average absolute error between the FNO model on case \textit{W-Nii-T}, \textit{W-Nii-T-SDF}, \textit{W-Nii-P} and \textit{W-Nii-P-SDF}.}
\label{Fig.DifferentTrainingMethods}
\end{figure}

This section focuses on two primary factors: (1) the impact of patch-based versus total-field training and (2) the inclusion or exclusion of the signed distance function (SDF) as an auxiliary feature. 
Fig.\ref{Fig.T5-1Visual} presents the velocity predictions of the FNO model for wind field forecasting under four training setups: \textit{W-Nii-T}, \textit{W-Nii-T-SDF}, \textit{W-Nii-P}, and \textit{W-Nii-P-SDF}. The results show small differences during the early stages of prediction, but as time progresses, the errors become more pronounced. As shown in Fig.\ref{Fig.DifferentTrainingMethods}, the \textit{W-Nii-P-SDF} setup demonstrates the best overall performance. Additionally, Table \ref{tab_RMS_T5-1} provides the maximum and mean RMS errors at time = 5 s for the different training setups, highlighting the impact of SDF inclusion. Notably, the \textit{W-Nii-P-SDF} case exhibits the best performance, with a maximum RMS error of 1.21 and a mean RMS error of 0.27, demonstrating the lowest error magnitude and highest consistency compared to the other methods. These results highlight the effectiveness of combining patch-based training with SDF data to enhance the FNO model’s accuracy and stability over time. Note that patch-based training enables the FNO model to focus on smaller, localized regions, allowing it to capture detailed wind flow patterns and reduce errors as predictions progress. The inclusion of SDF data further enhances performance by providing critical geometric information about the layout and structure of obstacles, improving the FNO model’s ability to predict wind dynamics in complex urban environments. By contrast, setups trained on the entire domain (\textit{W-Nii-T} and \textit{W-Nii-T-SDF}) can
capture global wind patterns but struggle with finer details, leading to faster error accumulation over time. These findings emphasize the importance of combining localized training strategies with geometric data to improve the accuracy and reliability of models for urban wind field forecasting, particularly for long-term simulations. 


\begin{table}[H]
\centering
\caption{Maximum and mean RMS errors of different training approaches and SDF impact at time = 1 s.}

\fontsize{0.8pt}{1.0pt}\selectfont

\setlength{\heavyrulewidth}{0.03mm}   
\setlength{\lightrulewidth}{0.015mm}  
\setlength{\aboverulesep}{0.15pt}     
\setlength{\belowrulesep}{0.15pt}     
\setlength{\arrayrulewidth}{0.1mm}   

\resizebox{\textwidth}{!}{
\begin{tabular}{c@{\extracolsep{5pt}}c@{\extracolsep{5pt}}c}
\toprule
\multicolumn{3}{c}{\textbf{Maximum and mean RMS errors at time = 5 [s]}} \\
\midrule
\textbf{Test Case} & \textbf{Maximum RMS Error} & \textbf{Mean RMS Error} \\
\midrule
\textit{W-Nii-P}  & 2.19 & 0.30 \\
\textit{W-Nii-T}  & 2.43 & 0.33 \\
\midrule
\textit{W-Nii-P-SDF} & 1.21 & 0.27 \\
\textit{W-Nii-T-SDF} & 2.23 & 0.28 \\
\bottomrule
\end{tabular}
}

\label{tab_RMS_T5-1}
\end{table}

\begin{table}
\centering
\caption{Radial energy spectrum differences of different training approaches and SDF impact at selected wave numbers.}
\fontsize{0.8pt}{1.0pt}\selectfont

\setlength{\heavyrulewidth}{0.03mm}   
\setlength{\lightrulewidth}{0.015mm}  
\setlength{\aboverulesep}{0.2pt}     
\setlength{\belowrulesep}{0.2pt}     
\setlength{\arrayrulewidth}{0.1mm}   

\resizebox{\textwidth}{!}{
\begin{tabular}{c@{\extracolsep{5pt}}c@{\extracolsep{5pt}}c}  

\toprule
\multicolumn{3}{c}{\textbf{Radial energy spectrum absolute differences at selected wave numbers}} \\
\midrule
\textbf{Wave Number} & \textit{W-Nii-P} & \textit{W-Nii-T} \\
\midrule
32        & 164.72      & 140.06       \\
64        & 51.07     &  53.11        \\
96        & 32.85     & 36.72        \\
128       & 27.96     & 34.20        \\
\midrule
\textbf{Wave Number} & \textit{W-Nii-P-SDF} & \textit{W-Nii-T-SDF} \\
\midrule
32          & 105.68   & 144.39      \\
64          & 41.61    & 51.59       \\
96          & 25.58    & 33.14       \\
128         & 21.29    & 27.18       \\
\bottomrule
\end{tabular}
}
\label{tab_Radial_errors_T5-1}
\end{table}

To further quantify the role of SDF data in capturing wind field features across different scales, we analyze the absolute error in the radial energy spectrum, as shown in Table \ref{tab_Radial_errors_T5-1}. The results reveal that the \textit{W-Nii-P-SDF} configuration consistently achieves lower absolute errors across all wave numbers, with particularly strong performance at a wave number of 32, corresponding to large-scale structures. At this wave number, the \textit{W-Nii-P-SDF} setup records an absolute error of 105.68, significantly lower than any other configuration. From the perspective of the Fourier energy spectrum, the combination of patch-based training and SDF data significantly enhances the FNO model’s adaptability across multiple scales. With the same hardware resource consumption, patch-based training allows the FNO model to capture a broader range of frequency features from the input data, effectively optimizing the distribution of predicted wind speed across all frequencies, particularly in the low-frequency (large-scale) range. This enables the FNO model to better capture global wind field patterns. Simultaneously, the inclusion of SDF data provides critical boundary information related to urban geometry, improving the FNO model’s ability to resolve high-frequency (local) variations. The synergy between these two approaches results in a substantial reduction in energy errors across the entire frequency spectrum.

\subsection{Evaluation of generalization across different wind directions}

\begin{figure}
\centering

\begin{minipage}{0.32\textwidth}
    \centering
    \begin{picture}(-50,0)
        \put(-60,122){\makebox(0,0)[lt]{(a)}}
    \end{picture}
    \includegraphics[scale=0.03]{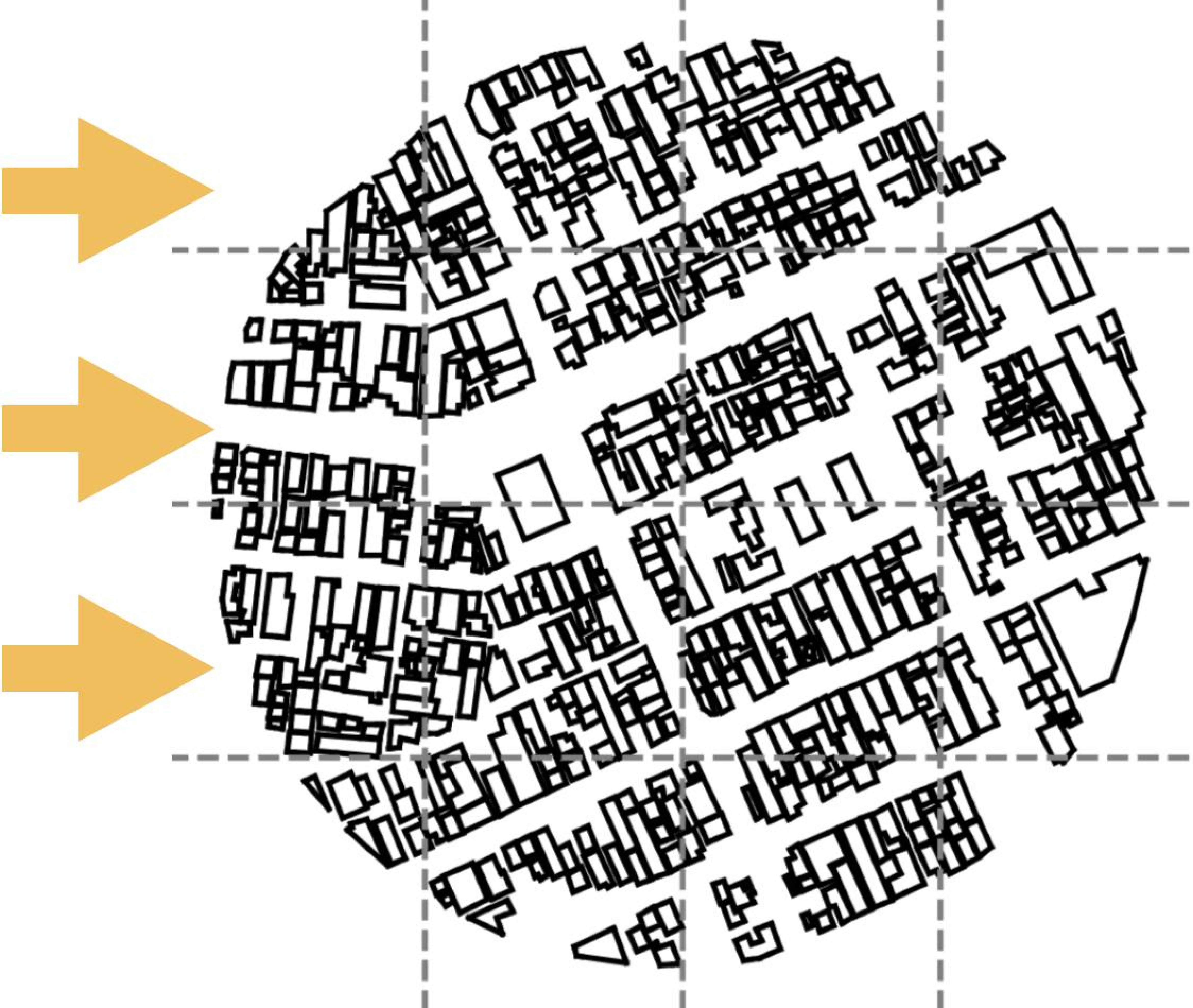}
\end{minipage}%
\hspace{0.05\textwidth}
\begin{minipage}{0.32\textwidth}
    \centering
    \begin{picture}(0,0)
        \put(-30,133){\makebox(0,0)[lt]{(b)}}
    \end{picture}
    \includegraphics[scale=0.25]{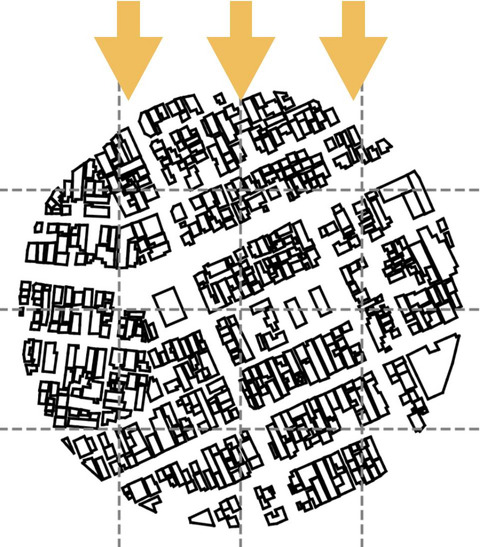}
\end{minipage}%
\hspace{0.02\textwidth}
\begin{minipage}{0.32\textwidth}
    \centering
    \begin{picture}(0,0)
        \put(-17,120){\makebox(0,0)[lt]{(c)}}
    \end{picture}
    \includegraphics[scale=0.25]{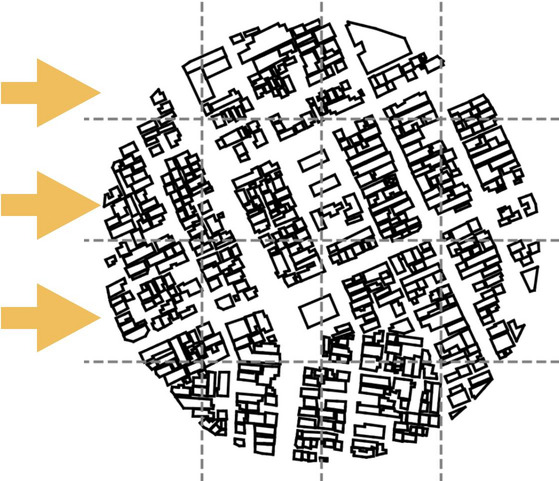}
\end{minipage}

\vspace{0.02\textwidth}

\caption{Visualization of the flow direction and modification: (a) West direction (\textit{W-Nii-CFD}), (b) North direction (\textit{N-Nii-P-SDF-R}), (c) North direction Rotated 90$^\circ$ counterclockwise (\textit{N-Nii-P-SDF-R}).}
\label{Fig.DifferentCitiesUrbanPattern2}
\end{figure}
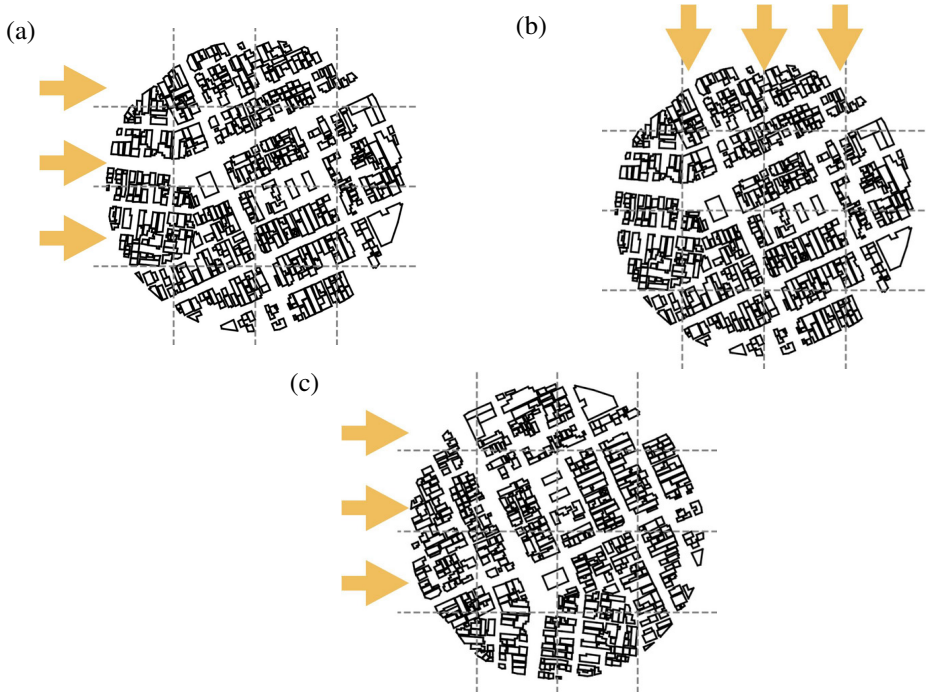

In addition to FNO's performance based on different training methods, in this section, we evaluate the generalization capability of the FNO model across varying wind directions. We designed two test scenarios. First, as illustrated in Fig.\ref{Fig.DifferentCitiesUrbanPattern2}, we used the \textit{N-Nii-P-SDF} dataset (Fig.\ref{Fig.DifferentCitiesUrbanPattern2}(a)) to test the FNO model’s performance on an unseen north wind direction. The results revealed that the FNO model struggled to generalize effectively to the north wind, highlighting its limitations in adapting to entirely new wind directions. To address this limitation, we rotated the north wind data to align with the west wind direction, which matches the orientation used in the training dataset. This rotated dataset, referred to as \textit{N-Nii-P-SDF-R} (Fig.\ref{Fig.DifferentCitiesUrbanPattern2}(b)), allowed us to evaluate the FNO model’s ability to adapt to changes in wind orientation while staying within the scope of familiar directional alignment. The results showed improved performance with the rotated data, demonstrating that the FNO model can leverage its learned features more effectively when the wind direction is aligned with the training data.

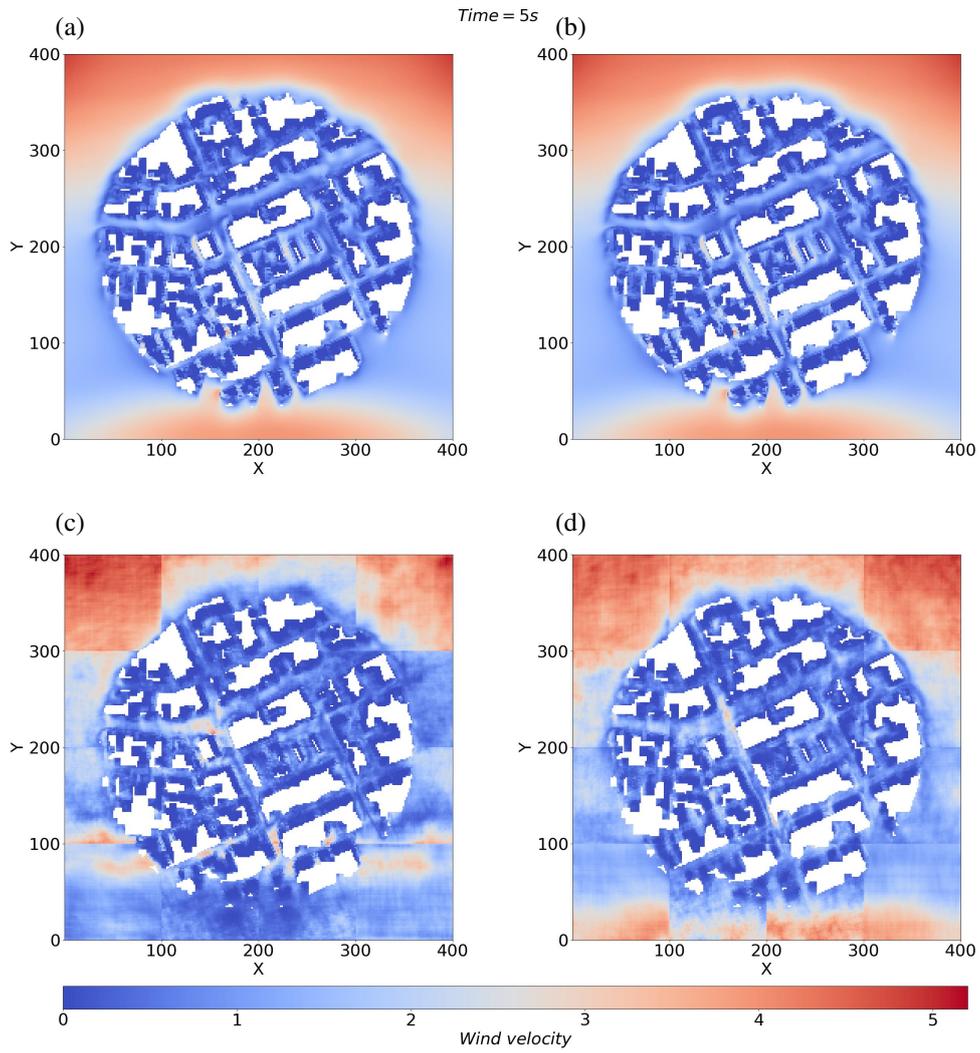
\begin{figure}
\centering
\includegraphics[scale=0.2]{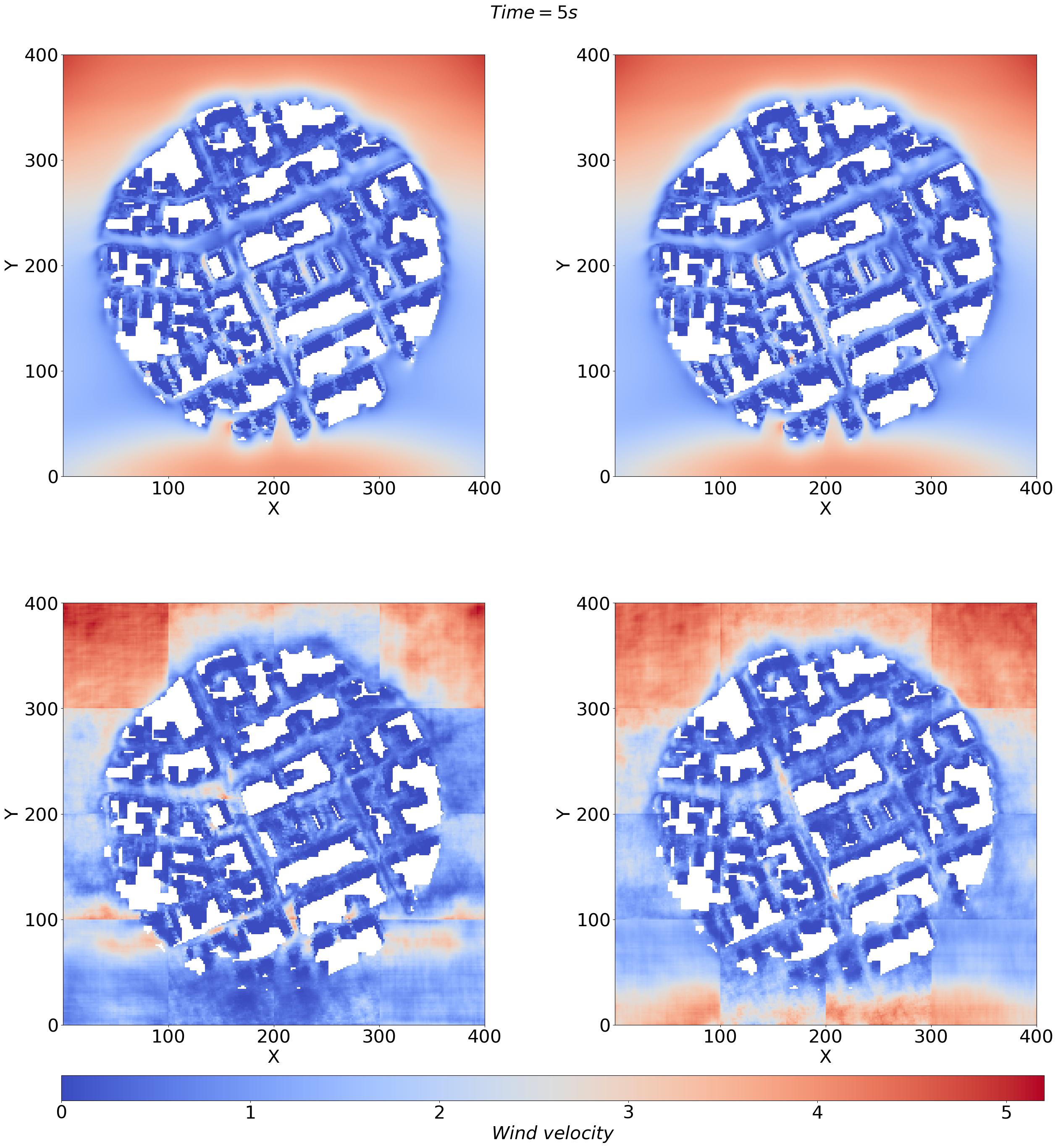}
\setlength{\unitlength}{1\textwidth} 
\begin{picture}(0,0)
    \put(-0.75,0.82){(a)}
    \put(-0.35,0.82){(b)}
    \put(-0.75,0.42){(c)}
    \put(-0.35,0.42){(d)}
\end{picture}
\caption{Visualization of the flow field outputs from FNO models for two cases: (a) and (b) \textit{N-Nii-P-CFD},  \textit{N-Nii-P-SDF}, (d) \textit{N-Nii-P-SDF-R}.}
\label{Fig.T5-2Visual}
\end{figure}

\begin{figure}
\centering
\includegraphics[scale=0.06]{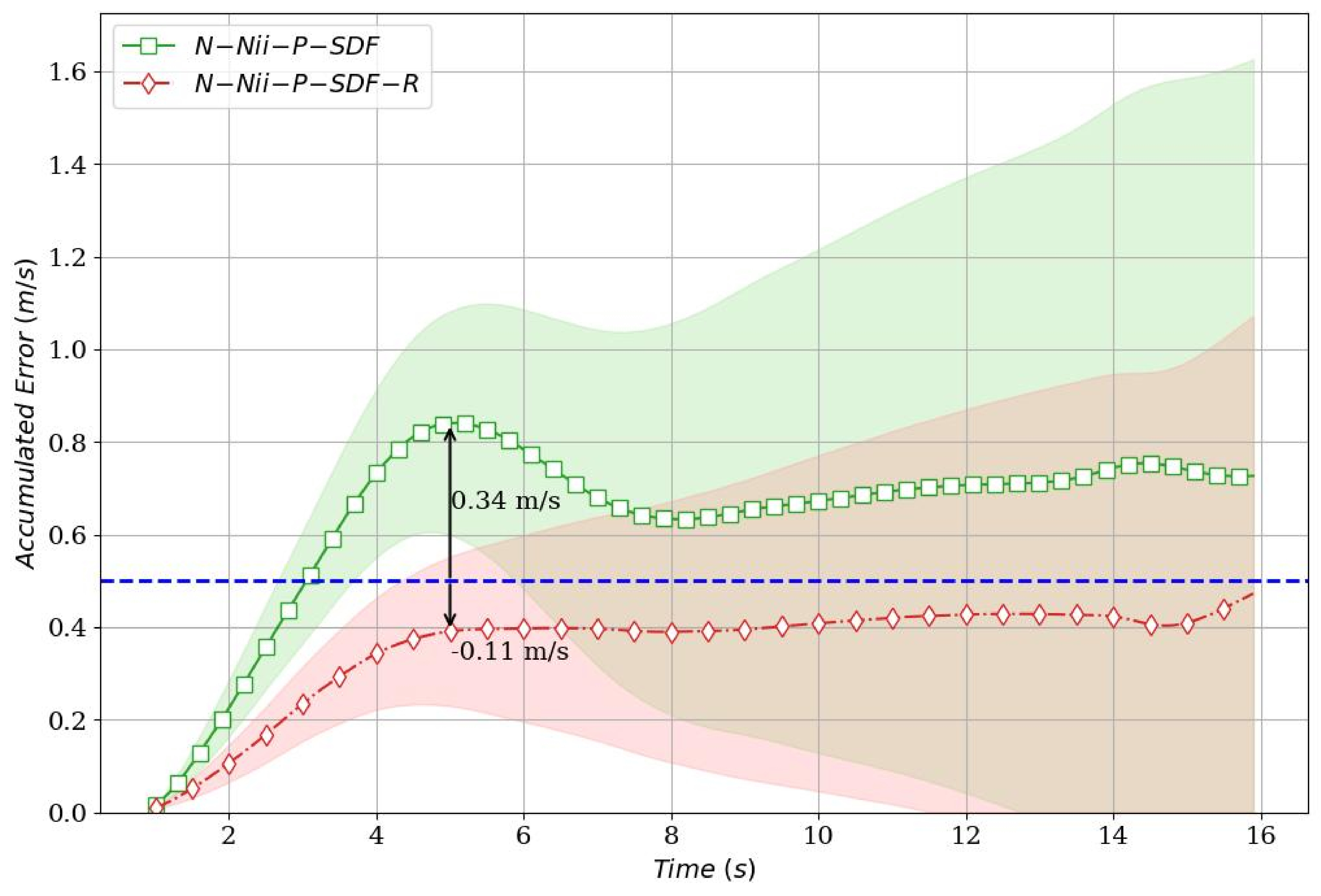}
\caption{Comparison of accumulated average absolute error between the FNO model on \textit{W-Nii-P} and \textit{W-Nii-P} case.}
\label{Fig.NorthRotateNoRotate1ineGraph}
\end{figure}

\begin{table}[b!]
\centering
\caption{Radial energy spectrum maximum and mean RMS errors of different training approaches and SDF impact at selected wave numbers.}
\fontsize{0.8pt}{1.0pt}\selectfont
\setlength{\heavyrulewidth}{0.03mm}   
\setlength{\lightrulewidth}{0.015mm}  
\setlength{\aboverulesep}{0.15pt}     
\setlength{\belowrulesep}{0.15pt}     
\setlength{\arrayrulewidth}{0.1mm}   
\resizebox{\textwidth}{!}{
\begin{tabular}{c@{\extracolsep{5pt}}c@{\extracolsep{5pt}}c}
\toprule
\multicolumn{3}{c}{\textbf{Maximum and mean RMS errors at time = 5 [s]}} \\
\midrule
\textbf{Test Case} & \textbf{Maximum RMS Error} & \textbf{Mean RMS Error} \\
\midrule
\textit{N-Nii-P-SDF}  & 4.73 & 0.65 \\
\textit{N-Nii-P-SDF-R}  & 4.72 & 0.34 \\
\bottomrule
\end{tabular}
}

\label{tab_Radial_errors_T5-2}
\end{table}

Firstly, we need to compare FNO's performance between the \textit{N-Nii-P-SDF} case and the \textit{N-Nii-P-SDF-R} case. Fig.\ref{Fig.NorthRotateNoRotate1ineGraph} and Fig.\ref{Fig.T5-2Visual} compare the accumulated average absolute error over time for different testing scenarios. The results indicate that the rotated scenario performs better, with error stabilizing earlier and consistently remaining lower throughout the simulation. At time = 5 s, the rotated scenario’s error is approximately 0.11 m/s below the accepTable CFD error level, while the original north wind scenario’s error is about 0.34 m/s above the baseline. Table \ref{tab_Radial_errors_T5-2} shows the maximum and mean RMS errors at time = 5 s for different training approaches and the impact of SDF at selected wave numbers. The results indicate that the \textit{N-Nii-P-SDF} case has a higher maximum RMS error (4.73) and mean RMS error (0.65) compared to the \textit{N-Nii-P-SDF-R} case, which has a maximum RMS error of 4.72 and a mean RMS error of 0.34. These findings suggest that aligning the test wind direction with the training conditions improves predictive accuracy. In the \textit{N-Nii-P-SDF} case, the mismatch between the wind direction and the training data indicates that the FNO model has not learned the spatial pattern of the wind field, leading to poorer predictive performance. In contrast, the \textit{N-Nii-P-SDF-R} case, where the wind field’s spatial pattern aligns with the training data, enables the FNO model to generalize more effectively. This alignment reduces prediction errors, enhances the robustness of urban wind field simulations, and significantly improves the FNO model’s predictive performance.

\begin{table}
\centering
\caption{Accumulated error and radial energy spectrum differences for different wind directions in Niigata.}

\fontsize{0.8pt}{1.0pt}\selectfont

\setlength{\heavyrulewidth}{0.03mm}   
\setlength{\lightrulewidth}{0.015mm}  
\setlength{\aboverulesep}{0.15pt}     
\setlength{\belowrulesep}{0.15pt}     
\setlength{\arrayrulewidth}{0.015mm}   

\resizebox{\textwidth}{!}{
\begin{tabular}{c@{\extracolsep{5pt}}c@{\extracolsep{5pt}}c}
\toprule
\multicolumn{3}{c}{\textbf{Radial Energy Spectrum Absolute Differences at Selected Wave Numbers}} \\
\midrule
Wave Number & \textit{N-Nii-P-SDF-R} & \textit{N-Nii-P-SDF} \\
\midrule
32  & 105.68  & 117.36 \\
64  & 41.61   & 44.31  \\
96  & 25.58   & 27.57  \\
128 & 21.29   & 23.03  \\
\bottomrule
\end{tabular}
}

\label{tab:combined_accumulated_error_spectrum_directions}
\end{table}

To explain the difference with respect to the Fourier frequency aspect, Table \ref{tab:combined_accumulated_error_spectrum_directions} shows absolute differences in the radial energy spectrum at various wave numbers (32, 64, 96, and 128). The rotated scenario exhibits smaller spectral differences across all wave numbers compared to the unrotated scenario, especially in the lower wave number range. From the perspective of spectral analysis, wind direction alignment significantly reduces the spectral differences. Specifically, the rotated wind field data exhibits smaller radial energy spectrum differences compared to the unrotated data, especially in the lower wave number range. Lower wave numbers represent large-scale variations in the wind field, and the FNO model is better able to predict these features without the interference of wind direction changes. These results indicate that aligning the test wind direction with the training wind direction allows the FNO model to more accurately capture the low-frequency (large-scale) energy of the wind field, thereby improving the prediction accuracy of the FNO model. This is because the rotated wind field becomes more consistent with the spatial patterns of the training data, enabling the FNO model to more effectively capture and retain frequency-domain features associated with large-scale wind field variations. This has practical implications for applying FNO to predict data under different wind directions. By rotating data from the same city under different wind directions to align with the training wind direction, optimal predictive results can be achieved.

\subsection{Evaluation of generalization ability across different urban geometries}

\begin{figure}
\centering
\begin{minipage}{0.32\textwidth}
    \centering
    \begin{picture}(-50,0)
        \put(-60,170){\makebox(0,0)[lt]{(a)}}
    \end{picture}
    \includegraphics[scale=0.5]{WindEng-elsarticle-V1/figures/niigata2dGridPatchesTraining.jpg}
\end{minipage}%
\hspace{0.05\textwidth}
\begin{minipage}{0.32\textwidth}
    \centering
    \begin{picture}(0,0)
        \put(-90,5){\makebox(0,0)[lt]{(b)}}
    \end{picture}
    \includegraphics[scale=0.5]{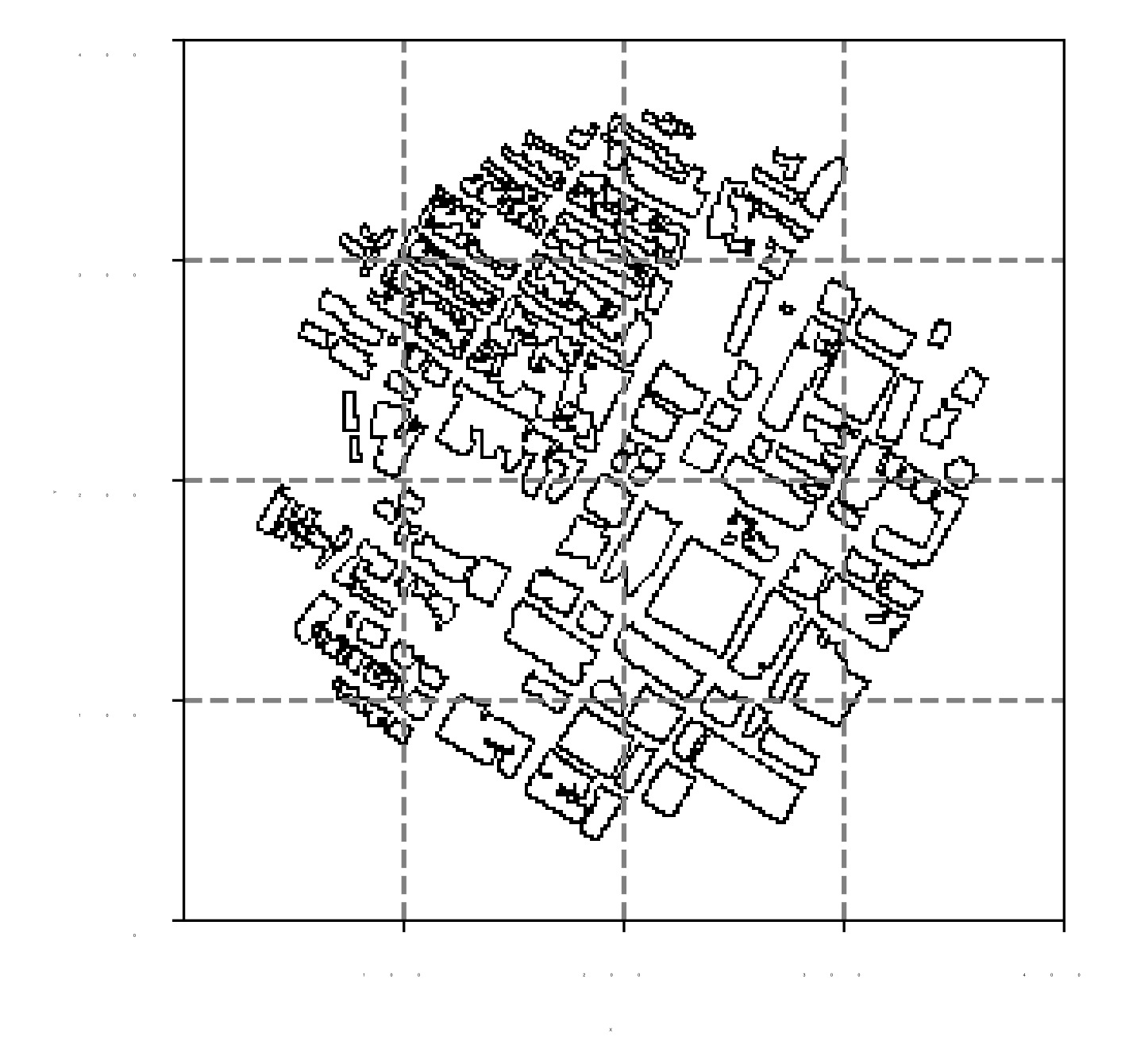}
\end{minipage}%
\hspace{0.02\textwidth}
\begin{minipage}{0.32\textwidth}
    \centering
    \begin{picture}(0,0)
        \put(-85,-5){\makebox(0,0)[lt]{(c)}}
    \end{picture}
    \includegraphics[scale=0.12]{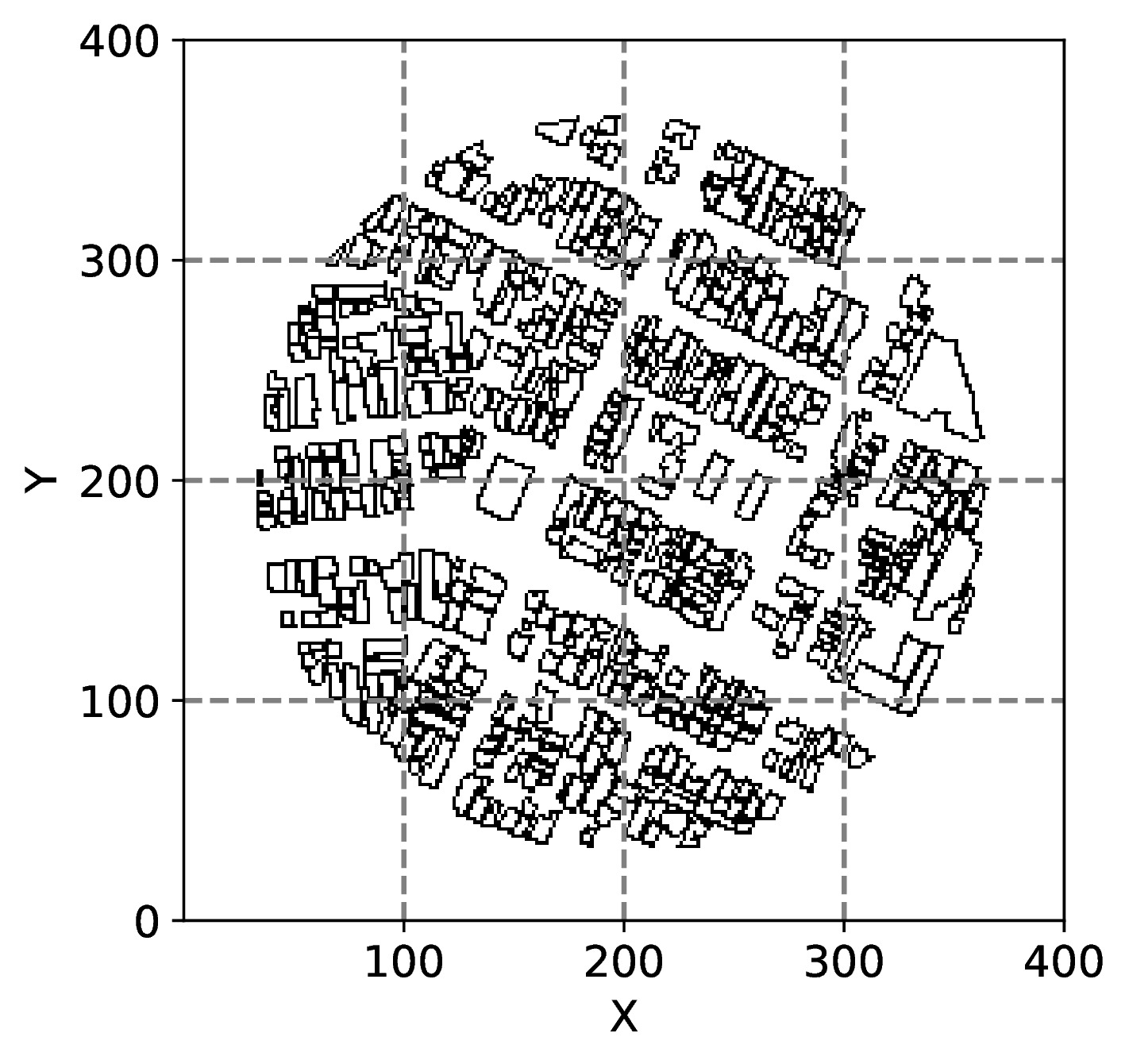}
\end{minipage}

\vspace{0.02\textwidth}
\caption{Visualization of the different city layout: (a) Niigata city geometry (\textit{W-Nii-P-SDF}). (b) Montreal city geometry (\textit{W-Mon-P-SDF}). (c) Vertically flipped Niigata city geometry (\textit{W-Nii-P-SDF-VF}).}
\label{Fig.DifferentCitiesUrbanPattern}
\end{figure}

After demonstrating FNO's performance on different wind directions in the same city, in this section, we evaluate FNO's performance on different cities' geometry, therefore, as displayed in Fig.\ref{Fig.DifferentCitiesUrbanPattern}, it examines the FNO model across three distinct urban geometries to analyze the FNO model's performance across different urban structures. Fig.\ref{Fig.DifferentCitiesUrbanPattern}(a) shows the Niigata layout, characterized by its dense and compact urban structure, which serves as the baseline for model training. Fig.\ref{Fig.DifferentCitiesUrbanPattern}(b) introduces a vertically flipped version of the Niigata layout, designed to test the FNO model’s sensitivity to mirrored configurations of familiar geometries. Fig.\ref{Fig.DifferentCitiesUrbanPattern}(c) presents the Montreal layout, featuring a markedly different urban structure, challenging the FNO model to generalize to new and unseen city configurations.

\begin{figure}
\centering
\includegraphics[scale=0.2]{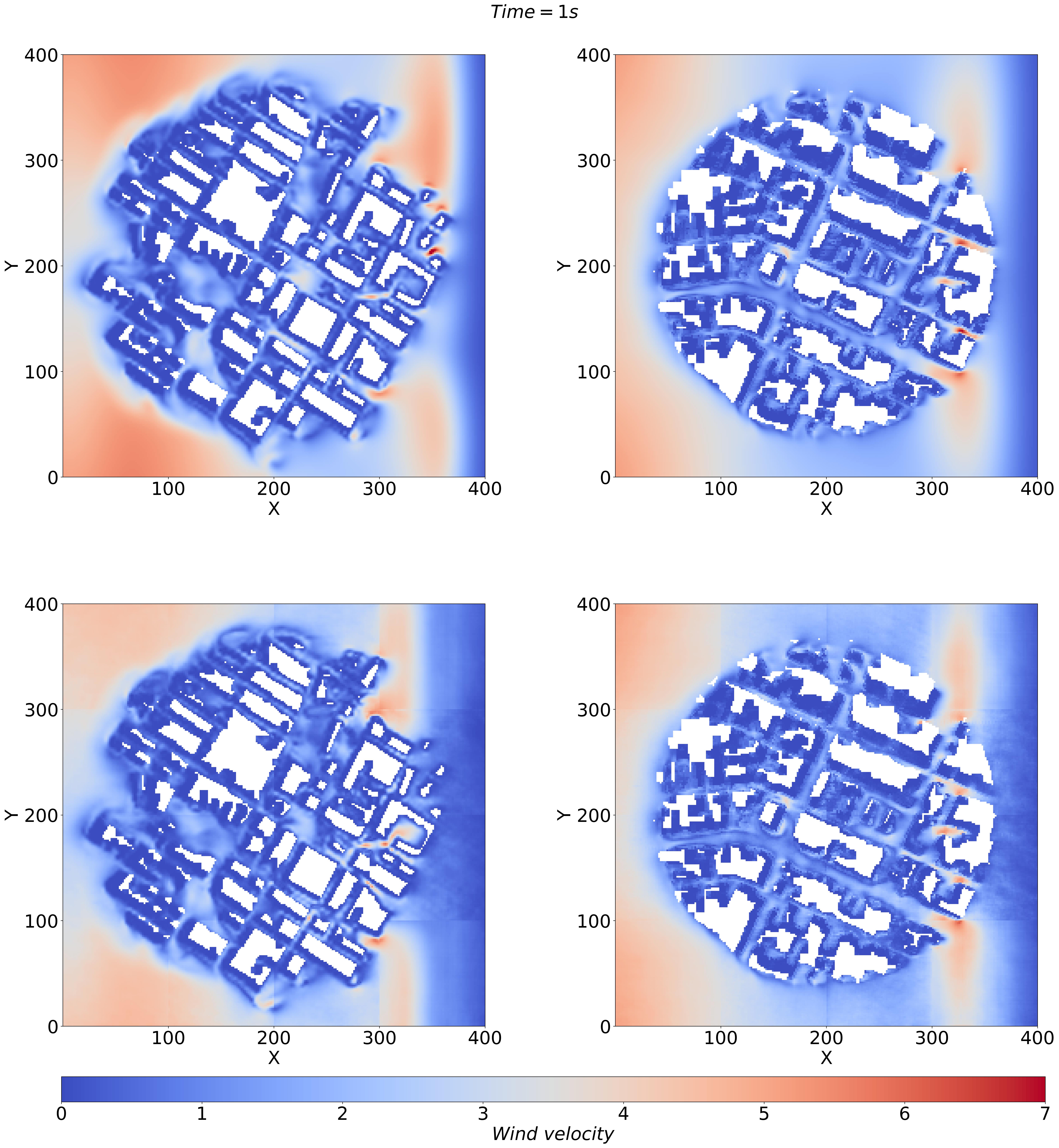}

\setlength{\unitlength}{1\textwidth} 
\begin{picture}(0,0)
    \put(-0.35,0.84){(a)}
    \put(0.05,0.84){(b)}
    \put(-0.35,0.44){(c)}
    \put(0.05,0.44){(d)}
\end{picture}
\caption{Visualization of the flow field outputs from FNO models for different urban layouts: (a) case \textit{W-Mon-P-CFD}, (b) case \textit{W-Nii-P-CFD}, (c) case \textit{W-Mon-P-SDF}, (d) case \textit{W-Nii-P-SDF-VF}.}
\label{Fig.5-3Visual}
\end{figure}

\begin{figure}
\centering
\includegraphics[scale=0.3]{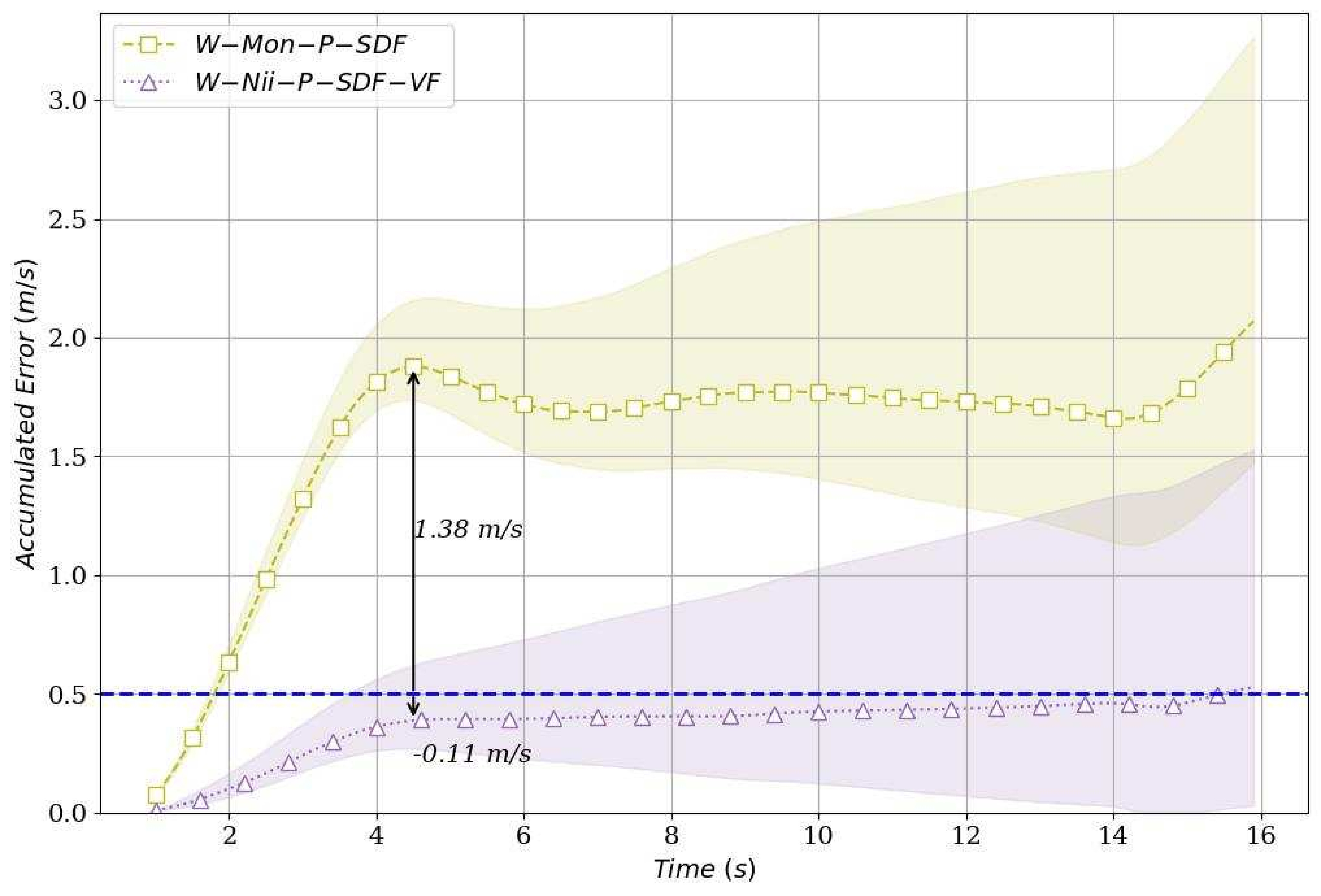}
\caption{Accumulated average absolute error over time for \textit{W-Mon-P-SDF} case and \textit{W-Nii-P-SDF-VF} case.}
\label{Fig.NorthRotateMontrealLineGraph111}
\end{figure}

The velocity predictions for the \textit{W-Mon-P-SDF} and \textit{W-Nii-P-SDF-VF} cases at time = 1s are visualized in figure \ref{Fig.5-3Visual}, providing deeper insights into the predictive performance of the FNO model across varying urban geometries. In the \textit{W-Mon-P-SDF} case, the Montreal layout, characterized by a significantly different and less compact urban structure compared to the training data, is analyzed. In contrast, the \textit{W-Nii-P-SDF-VF} case examines the vertically flipped Niigata layout to assess the FNO model’s ability to adapt to mirrored configurations of familiar geometries. Clear differences in the FNO model’s performance across these scenarios are revealed through the visualizations. For the Montreal layout (\textit{W-Mon-P-SDF}), more pronounced discrepancies are observed near structural boundaries, such as around buildings and open spaces, which highlight challenges in generalizing to geometries that differ significantly from the training data. Higher error accumulation in regions with complex wind interactions is particularly evident, indicating limitations in the FNO model’s adaptability to entirely new layouts. In contrast, higher accuracy and smoother transitions across the velocity field are observed for the flipped Niigata layout (\textit{W-Nii-P-SDF-VF}), suggesting that the structural similarity to the training data allows the FNO model to perform more effectively.

The FNO model's performance across different urban geometries is shown in figure \ref{Fig.NorthRotateMontrealLineGraph111}. The \textit{W-Mon-P-SDF} case exhibits a significantly higher accumulated error, reaching approximately 1.26 m/s, surpassing the accepTable CFD error threshold at time = 15 s. This highlights the FNO model’s difficulty in adapting to the distinct urban structure of Montreal. In contrast, the \textit{W-Nii-P-SDF-VF} case shows an error progression similar to that of the original Niigata City scenario, stabilizing at around -0.09 m/s. the FNO model’s poor performance on the Montreal layout (the \textit{W-Mon-P-SDF} case) compared to its better accuracy on the vertically flipped Niigata layout can be explained by several factors. Firstly, significant differences exist between the urban structures of Montreal and Niigata, particularly in terms of building density, road layout, and overall city geometry. Since the FNO model was trained on Niigata west wind field data, the structural features of the training data differ substantially from those of Montreal. Consequently, when the FNO model encounters Montreal test data (the \textit{W-Mon-P-SDF} case), it struggles to effectively generalize the spatial features learned during training to the new urban structure, leading to higher accumulated errors. This underscores the FNO model's limitations in generalizing to cities with significantly different structural characteristics. The vertically flipped Niigata layout, despite being mirrored, retains geometric features similar to the training data. This structural resemblance allows the FNO model to better adapt to the changes and maintain higher prediction accuracy. This indicates that the FNO model is more effective in scenarios where structural variations are minimal, particularly when the urban structure remains closely aligned with the training data. In such cases, the FNO model can more effectively apply the spatial features and wind field patterns it learned during training to new scenarios, resulting in more accurate predictions.

\begin{table}
\centering
\caption{Radial energy spectrum maximum and mean RMS errors of different training approaches and SDF impact at selected wave numbers.}

\fontsize{0.8pt}{1.0pt}\selectfont

\setlength{\heavyrulewidth}{0.03mm}   
\setlength{\lightrulewidth}{0.015mm}  
\setlength{\aboverulesep}{0.15pt}     
\setlength{\belowrulesep}{0.15pt}     
\setlength{\arrayrulewidth}{0.1mm}   

\resizebox{\textwidth}{!}{
\begin{tabular}{c@{\extracolsep{5pt}}c@{\extracolsep{5pt}}c}
\toprule
\multicolumn{3}{c}{\textbf{Maximum and mean RMS errors at time = 1 [s]}} \\
\midrule
\textbf{Test Case} & \textbf{Maximum RMS Error} & \textbf{Mean RMS Error} \\
\midrule
\textit{W-Mon-P-SDF}  & 5.97 & 0.40 \\
\textit{W-Nii-P-SDF-VF}  & 2.51 & 0.06 \\
\bottomrule
\end{tabular}

}

\label{tab_Radial_errors_T5-3}
\end{table}

\begin{table}
\centering
\caption{Comparison of accumulated error, radial energy spectrum differences, and SSIM for different urban geometries.}

\resizebox{\textwidth}{!}{%

\fontsize{0.8pt}{1.0pt}\selectfont

\setlength{\heavyrulewidth}{0.025mm}   
\setlength{\lightrulewidth}{0.0125mm}  
\setlength{\aboverulesep}{0.1pt}     
\setlength{\belowrulesep}{0.1pt}     
\setlength{\arrayrulewidth}{0.0125mm}

\begin{tabular}{c@{\extracolsep{5pt}}c@{\extracolsep{5pt}}c}
\toprule
\multicolumn{3}{c}{\textbf{Radial Energy Spectrum Absolute Differences at Selected Wave Numbers}} \\
\midrule
Wave Number & \textit{W-Nii-P-SDF-VF} & \textit{W-Mon-P-SDF} \\
\midrule
32  & 107.26 & 309.22 \\
64  & 40.61  & 90.61  \\
96  & 24.07  & 35.17  \\
128 & 20.49  & 21.29  \\
\midrule
\multicolumn{3}{c}{SSIM Comparison for Urban Geometries} \\
\midrule
\textbf{Urban Geometry} & \multicolumn{2}{c}{\textbf{SSIM}} \\
\midrule
       \textit{W-Nii-P-SDF-VF} & \multicolumn{2}{c}{0.5784} \\
       \textit{W-Mon-P-SDF}    & \multicolumn{2}{c}{0.4816} \\
\bottomrule
\end{tabular}%
}
\label{tab:combined_comparison_models-diffcities}
\end{table}

As shown in Table \ref{tab:combined_comparison_models-diffcities}, For the \textit{W-Nii-P-SDF-VF} case, the energy spectrum differences are significantly lower, particularly at lower wavenumbers, with a value of 107.26 at a wavenumber of 32. This indicates that when the urban geometry of the test data is more closely aligned with the training data, the FNO model is better able to capture the large-scale energy distribution of the wind field and can effectively leverage the spatial features it has learned, resulting in more accurate large-scale wind field predictions. Conversely, the Montreal urban layout, with its significant structural differences from Niigata, leads to greater discrepancies in the wind field's energy patterns. These differences make it more challenging for the FNO model to represent large-scale features accurately, resulting in higher energy spectrum differences. This finding underscores the possible importance of geometric consistency between the training and testing urban layouts for reliable predictions of large-scale energy patterns in wind field simulations. To further analyze the geometric differences theoretically, SSIM values \cite{bakurov2022structural} provide additional insights into the structural similarity between different urban layouts. The SSIM value of the \textit{W-Mon-P-SDF} case is 0.5784, while the SSIM value of the \textit{W-Nii-P-SDF-VF} case is notably lower at 0.4816. The higher SSIM value for the \textit{W-Nii-P-SDF-VF} case reflects a closer structural resemblance to the original Niigata layout, which helps the FNO model maintain similar prediction accuracy. This is reflected in the lower accumulated errors and smaller energy spectrum differences observed in he \textit{W-Nii-P-SDF-VF} case. In contrast, the lower SSIM value of the \textit{W-Mon-P-SDF} case signifies a greater structural dissimilarity between the two cities, resulting in larger prediction errors and more pronounced energy spectrum discrepancies. These results demonstrate that structural similarity, as measured by SSIM, is a critical factor in determining the FNO model’s ability to generalize effectively across different urban layouts, particularly for accurate representation of large-scale wind field patterns.
This finding related to SSIM further highlights the practical importance of testing the FNO model on urban layouts with consistent geometric structures. When the FNO model is trained on data with geometries similar to the target city, such as Niigata and its flipped geometry, it demonstrates better generalization, resulting in more accurate wind field predictions. Conversely, when applied to cities with significantly different geometries, such as Montreal, the FNO model struggles to adapt its learned features to the new structure, leading to reduced prediction accuracy.

\section{Discussions}
\label{discuss}
The deep learning-based FNO model proposed in this study for urban wind farm simulations is primarily data-driven and has not fully leveraged the a priori knowledge inherent in traditional CFD methods. While the FNO model demonstrates excellent prediction efficiency, it lacks the mathematical foundation and interpretability of CFD models. A promising direction for future research is to integrate the mathematical principles of CFD into the FNO framework. By incorporating physical loss functions, the training and testing processes could better align with the physical laws of fluid dynamics, enhancing the FNO model’s physical interpretability. At the same time, this approach would retain the FNO’s ability to extract global features through Fourier transforms, allowing the FNO model to capture the global dynamics of the wind field while maintaining a high degree of physical consistency. Further improvements are also required to integrate the FNO model with the CityFFD solver. Currently, the two systems function independently, but future work could explore their combination to further accelerate the simulation process. Such integration would enable the FNO model to reduce computation time while preserving high accuracy, effectively complementing traditional CFD simulations.

Our experiments indicate that the FNO model tends to accumulate errors beyond the accepTable CFD threshold of 0.5 m/s during the later stages of prediction (e.g., after 15 seconds). Although the FNO model performs well initially, its accuracy diminishes over time. To enhance long-term prediction capabilities, future efforts should focus on increasing the size of the training dataset, optimizing the FNO model architecture, and incorporating more diverse urban wind field data. These steps would improve the FNO model’s stability, reduce error accumulation, and enhance its performance over longer prediction intervals. Additionally, the current study is limited to wind field data from Niigata and Montreal. To develop a more generalizable model, future research should expand the dataset to include wind field data from a wider range of cities, encompassing more complex urban geometries and diverse climatic conditions. This broader dataset would refine the FNO model’s ability to generalize across different urban environments, making it more robust and versatile in practical applications.

\section{Conclusions}
\label{conclu}

In this paper, a series of in-depth learning experiments were carried out to evaluate the generalization capability of the FNO model in urban wind field simulation tasks. These experiments were designed to assess the FNO model's performance under varying wind directions and across diverse urban layouts, with all scenarios implemented within the FNO framework. Based on the results of these experiments, the following conclusions can be drawn:

\textbf{Effectiveness of patch-based training and SDF information}: Due to GPU hardware constraints, training the FNO model on entire wind fields is challenging, making patch-based training a more efficient strategy. This approach mitigates hardware limitations while capturing localized dynamic features with greater precision and significantly reducing computational costs. It offers a practical balance between accuracy and efficiency, particularly in resource-limited environments. Additionally, without Signed Distance Function (SDF) data, the FNO model struggles to accurately predict urban wind fields, especially in dense and complex regions where physical boundaries introduce significant disruptions. Incorporating SDF data provides critical spatial information about buildings and obstacles, enhancing the FNO model’s ability to perceive physical boundaries and yielding more accurate and realistic predictions. This improvement is especially valuable for applications such as wind energy optimization and urban air pollution management.


\textbf{Generalization to various wind directions:} Despite being trained solely on west wind data, the FNO model exhibited strong generalization capabilities when tested under north wind conditions. This indicates that the FNO model is capable of capturing fundamental dynamic features of the wind field, allowing it to adapt to wind directions that were not present in the training data. the FNO model’s ability to generalize from west wind to north wind conditions demonstrates its flexibility and utility, particularly in scenarios where training data for multiple wind directions is unavailable. This adaptability enhances the FNO model's applicability to a wide range of urban wind field prediction tasks.

\textbf{Generalization to different urban layouts:} The FNO model's performance in cross-city applications is significantly influenced by the structural similarity between the target city and the city used for training. The experiments showed that when the target city shares similar architectural layout and terrain characteristics with the training city—such as in the case of the flipped Niigata layout—the FNO model maintained a higher prediction accuracy. However, when there is a significant difference in structure, as seen with Montreal, the FNO model’s accuracy decreases notably. This suggests that structural similarity between cities is critical for successful cross-city wind field predictions, particularly in areas with dense buildings or complex terrain. To ensure accurate predictions in cross-city applications, it is essential to evaluate the physical characteristics of the target city against those of the training city, emphasizing the need for alignment in urban structure and geometry.




\section*{Declaration of competing interest}

The authors declare no competing financial or non-financial interests.

\section*{Data availability}

Other data supporting the findings of this study are available from the corresponding author upon reasonable request.

\section*{Code availability}

The codes used to generate the results of this study are available from the authors on a reasonable request.



\section*{Appendix A : FNO approach}

According
 to CNNs' capability at processing local features, it has been adopted also in deep learning tasks. However, they are relatively weak at capturing global information \cite{rawat2017deep}. This limitation arises from the reliance of CNNs on local convolutional kernels to extract edges and details in images, whereas the solutions to PDEs are continuous functions with strong global dependencies. Consequently, traditional methods struggle to efficiently handle PDEs.
To address this challenge, Li et al. proposed an innovative approach that parameterizes the neural network model in Fourier space to learn the Navier-Stokes equations rather than using the traditional Euclidean space framework \cite{li2020fourier}. This approach enables the FNO model to mimic the characteristics of pseudo-spectral methods, thereby enhancing its ability to approximate PDE solutions \cite{wang2024prediction,li2024geometry}.
Deep learning experimental results demonstrate that the FNO not only significantly improves the computational accuracy in approximating the Navier-Stokes equations but also surpasses the performance of existing methods, achieving unprecedented levels of precision \cite{li2020fourier}. More importantly, the FNO has the capability to learn the mapping between input parameters and the corresponding solutions from a finite set, thus enabling it to handle an entire family of PDEs rather than being limited to solving a single equation \cite{li2020fourier}.

Let \( \mathcal{D} \subset \mathbb{R}^n \) be a bounded open set, representing the spatial region under consideration. In this domain, we define the target nonlinear mapping \( \mathcal{G}^\dagger \), which maps from the function space \( \mathcal{X} \) to \( \mathcal{Y} \), i.e., \( \mathcal{G}^\dagger: \mathcal{X} \to \mathcal{Y} \). Here, \( \mathcal{X}(\mathcal{D}; \mathbb{R}^{d_x}) \) and \( \mathcal{Y}(\mathcal{D}; \mathbb{R}^{d_y}) \) are separable Banach spaces of functions defined on the region \( \mathcal{D} \), with values in \( \mathbb{R}^{d_x} \) and \( \mathbb{R}^{d_y} \), respectively, representing the value domains of the input and output functions. The choice of these spaces implies that the functions under consideration not only possess a well-defined mathematical structure (such as boundedness and additivity), but also maintain continuity and appropriate norm structures during the mapping process. Specifically, \( d_x \) and \( d_y \) represent the dimensionality of the input and output functions, determining the complexity of the mapping and the dimensionality of the target space. This formal setting provides a suitable mathematical framework for analyzing and understanding the properties of the mapping \( \mathcal{G}^\dagger \), particularly within the context of Banach space theory \cite{fabian2011banach}.  
To approximate the target nonlinear mapping \( \mathcal{G}^\dagger \), the Fourier Neural Operator (FNO) constructs a mapping \( \mathcal{G}: \mathcal{X} \times \Phi \to \mathcal{Y} \), where:
\( \mathcal{X} \) is a Banach space defined on a domain \( \mathcal{D} \subset \mathbb{R}^n \), representing the input function space. Specifically, \( \mathcal{X}(\mathcal{D}; \mathbb{R}^{d_x}) \) is the space of functions defined on \( \mathcal{D} \) and taking values in \( \mathbb{R}^{d_x} \), where \( d_x \) denotes the dimensionality of the input functions.
- \( \mathcal{Y} \) is also a Banach space, representing the output function space. In particular, \( \mathcal{Y}(\mathcal{D}; \mathbb{R}^{d_y}) \) is the space of functions taking values in \( \mathbb{R}^{d_y} \), where \( d_y \) denotes the dimensionality of the output functions.
- \( \Phi \) is a parameter space that contains all possible sets of parameters for parameterizing the mapping \( \mathcal{G} \). The mapping \( \mathcal{G} \) is parameterized by \( \varphi \in \Phi \), meaning that for each \( \varphi \), there is a corresponding mapping \( \mathcal{G}(\cdot, \varphi): \mathcal{X} \to \mathcal{Y} \).

In this framework, the optimal parameter \( \varphi^\dagger \in \Phi \) is determined through a purely data-driven approach, meaning that the mapping \( \mathcal{G}(\cdot, \varphi^\dagger) \) approximates the target mapping \( \mathcal{G}^\dagger \). Specifically, we aim to ensure that the parameterized mapping \( \mathcal{G}_{\varphi^\dagger} = \mathcal{G}(\cdot, \varphi^\dagger) \) closely approximates the target mapping \( \mathcal{G}^\dagger \), i.e., \( \mathcal{G}_{\varphi^\dagger} \approx \mathcal{G}^\dagger \). This is achieved through a data-driven optimization process, where the parameter optimization is based solely on the training data, without relying on prior physical knowledge.

The training process of FNO is structured as an iterative process, represented as \( \mu_0 \to \mu_1 \to \dots \to \mu_T \), where \( k = 0, 1, \dots, T-1 \) is an index sequence representing a series of functions \( \mu_k \). Each function \( \mu_k \) is a function defined on the domain \( \mathcal{D} \), and the output of each function is a vector in \( \mathbb{R}^{d_z} \), meaning that each function \( \mu_k \) outputs a \( d_z \)-dimensional vector. The purpose of this iterative structure is to progressively approximate the target mapping \( \mathcal{G}^\dagger \) by gradually updating the function sequence \( \mu_k \) through optimization in each step.

The overall architecture of the adapted FNO model is illustrated below in Fig.A.1:

\begin{figure}
\centering
\includegraphics[scale=0.5]{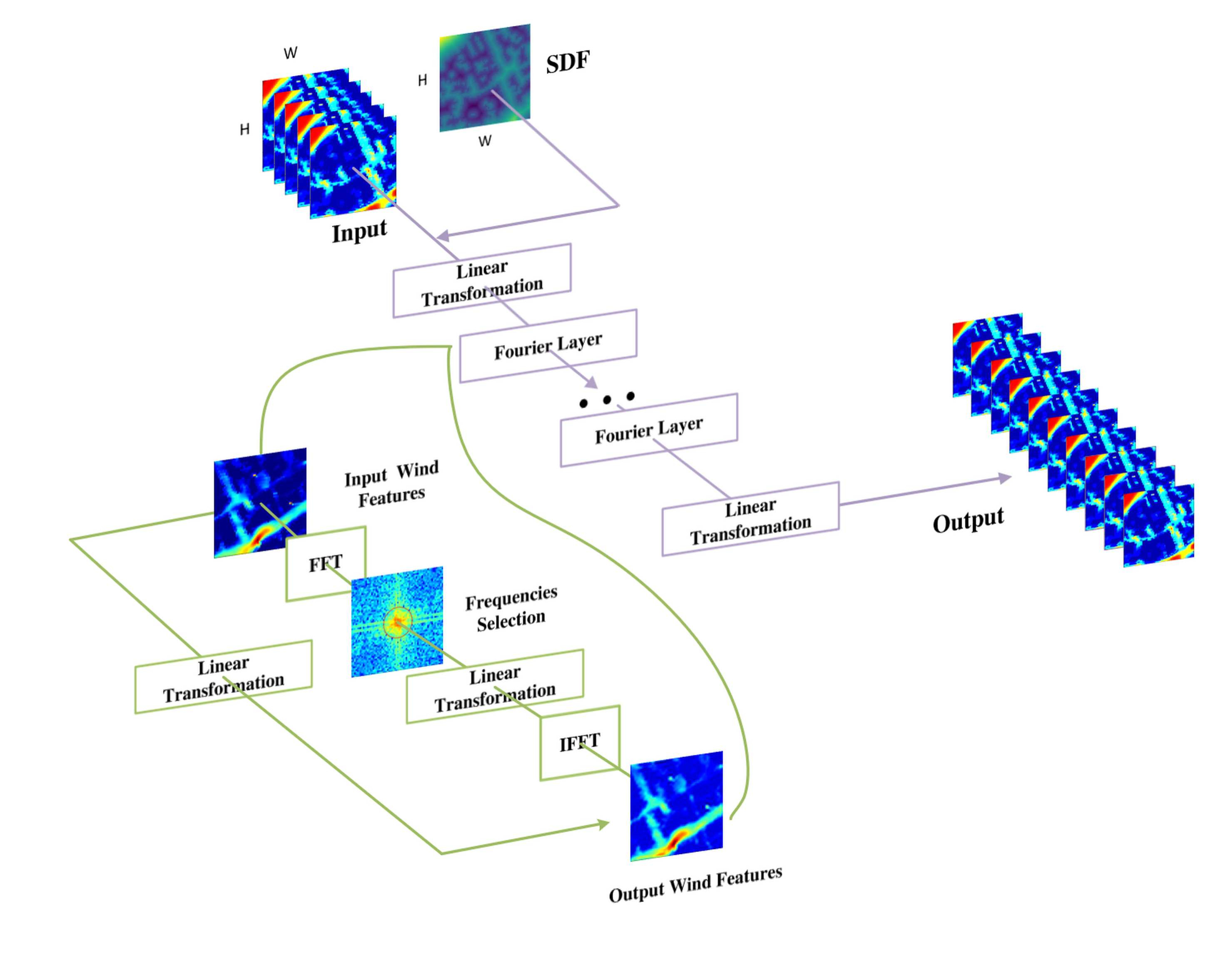}
\caption*{Fig. A.1 FNO architecture for enhanced prediction of urban wind flow environments.}
\label{Fig.fnomodel}
\end{figure}


Initially, the input function \( f(y) \) is mapped into a high-dimensional space through a fully connected layer \( P \). The purpose of this process is to map the input signal from its original space to a higher-dimensional space, allowing for a better capture of complex feature relationships. 
Subsequently, the representation undergoes processing via a series of Fourier layers, where the signal is transformed in the frequency domain to capture its features, while maintaining the tensor shape throughout the entire process. This ensures that the dimensionality and structure of the data remain unchanged during processing, thereby ensuring coherence and operability of the mapping process.
Finally, after processing through the Fourier layers, the output function \( v(y) \) is projected back to the target dimension via another fully connected layer \( T \). This step converts the high-dimensional representation into the final output, aligning with the requirements of the target task.
Within this framework, the high-dimensional representation is iteratively updated through a sequence of Fourier layers. Each layer transforms the input signal to enhance its feature representation, enabling the model to progressively approximate the target output. The iterative update process is described by Eq.\ref{eq:state_update}.

In this process, the mapping \( \mathcal{M}: \mathcal{X} \times \Theta_{\mathcal{M}} \to \mathcal{M} \left( \mathcal{V}(C; \mathbb{R}^{d_s}), \mathcal{V}(C; \mathbb{R}^{d_s}) \right) \) defines a bounded linear operator acting on the function space \( \mathcal{V}(C; \mathbb{R}^{d_s}) \). Here, \( \mathcal{X} \) represents the input space, \( \Theta_{\mathcal{M}} \) is the parameter space of the mapping \( \mathcal{M} \), and \( \mathcal{V}(C; \mathbb{R}^{d_s}) \) is the function space of \( \mathbb{R}^{d_s} \)-valued functions defined on the set \( C \). The parameter \( \xi \in \Theta_{\mathcal{M}} \) controls the specific form of the mapping \( \mathcal{M} \), thus influencing the transformation of the high-dimensional representation.
Additionally, the linear transformation \( N: \mathbb{R}^{d_s} \to \mathbb{R}^{d_s} \) is used to perform linear operations on the representation, changing its dimensionality or structure for subsequent processing. The nonlinear activation function \( \psi: \mathbb{R} \to \mathbb{R} \) introduces nonlinearity, enhancing the model's expressiveness and enabling it to learn complex feature mappings.
\begin{equation}
\label{eq:state_update}
w_{t+1}(y) = \varphi \left( M w_t(y) + \left( \mathcal{L}(b; \eta) w_t \right)(y) \right), \quad \forall y \in D
\end{equation}

In Eq.\ref{eq:state_update}, \( w_{t+1}(y) \) represents the updated value of the function at time step \( t+1 \). This update is composed of two parts: first, \( M w_t(y) \) is the result of applying the linear transformation \( M \) to the current state \( w_t(y) \); second, \( \mathcal{L}(b; \eta) w_t(y) \) is the outcome of applying a linear operator \( \mathcal{L}(b; \eta) \), parameterized by \( \eta \) and input \( b \), to \( w_t(y) \). Finally, the function \( \varphi \) represents a nonlinear activation function that processes the sum of these two terms.

Next, let \( \mathcal{G} \) and \( \mathcal{G}^{-1} \) represent the Fourier transform and its inverse, respectively, which map the input function from the spatial domain to the frequency domain. The Fourier transform of the function \( f \) is denoted by \( \mathcal{G}f \), and the inverse Fourier transform is denoted by \( \mathcal{G}^{-1} \). The Fourier integral operator is represented in Eq.\ref{eq:kernel_eta}, where \( T_\eta \) is the Fourier transform of the periodic function \( \lambda: \bar{D} \to \mathbb{R}^{d_w \times d_w} \), and \( \eta \) is the set of parameters defining the linear operator \( \mathcal{L}(\eta) \).

\begin{equation}
\label{eq:kernel_eta}
\left( \mathcal{L}(\eta)w_t \right)(y) = \mathcal{G}^{-1}\left( T_\eta \cdot \mathcal{G}w_t \right)(y), \quad \forall y \in D
\end{equation}

In Eq.\ref{eq:kernel_eta}, the operator \( \mathcal{L}(\eta) \) is a linear transformation defined by the parameter \( \eta \), applied to the function \( w_t(y) \) in the frequency domain. Specifically, \( \mathcal{G} \) is the Fourier transform, which maps \( w_t(y) \) to the frequency domain. \( T_\eta \) is the weight tensor that applies a convolution in the frequency domain. The convolution result is then mapped back to the spatial domain using the inverse Fourier transform \( \mathcal{G}^{-1} \).

The frequency modes \( \ell \in D \) can be expanded using a Fourier series, where \( \ell \in \mathbb{Z}^d \) represents the discrete frequency modes in the \( d \)-dimensional space. To achieve a finite-dimensional parameterization, the FNO truncates the Fourier series at the maximum number of modes \( \ell_{\text{max}} \), defined as Eq.\ref{eq6}:

\begin{equation}
\label{eq6}
\ell_{\text{max}} = |\Lambda_{\ell_{\text{max}}}| = \left| \ell \in \mathbb{Z}^d: |\ell_j| \leq \ell_{\text{max}, j}, \, j=1,\dots,d \right|
\end{equation}

Through truncation at \( \ell_{\text{max}} \), the FNO uses only a limited number of frequency modes for the computation.

Additionally, the domain \( D \) is discretized with \( m \) points, meaning that \( w_t \in \mathbb{R}^{m \times d_w} \) is the value of the function at the discretized points, and \( \mathcal{G}(w_t) \in \mathbb{C}^{m \times d_w} \) is the Fourier-transformed representation of \( w_t \) in the frequency domain. The weight tensor \( T_\eta \in \mathbb{C}_{\ell_{\text{max}} \times d_w \times d_w} \) contains the selected Fourier modes. By truncating higher-order modes, the representation becomes finite-dimensional.

Eq.\ref{eq:T_eta} describes the convolution of the Fourier-transformed signal with the weight tensor \( T_\eta \):

\begin{equation}
\label{eq:T_eta}
T_\eta \cdot \mathcal{G}w_t = \sum_{k=1}^{d_w} T_{\eta,k,l,k} \left( \mathcal{G}w_t \right)_{\ell,k}, \quad \ell = 1, \dots, \ell_{\text{max}}, \quad k = 1, \dots, d_w
\end{equation}

In Eq.\ref{eq:T_eta}, \( T_{\eta,k,l,k} \) represents the elements of the weight tensor \( T_\eta \), which is applied to the frequency modes in the Fourier-transformed signal \( \mathcal{G}w_t \). The sum over the modes \( \ell \) and the frequency channels \( k \) results in the updated representation of the signal.

In CFD modeling, it is common to discretize the flow uniformly with a resolution of \( p_1 \times \dots \times p_d = m \), and to replace \( \mathcal{G} \) with the Fast Fourier Transform (FFT). Let \( g \in \mathbb{R}^{m \times d_u} \) be a matrix of size \( m \times d_u \), where \( m \) represents the number of discretized grid points, and \( d_u \) denotes the number of channels (i.e., the dimensionality of the data). For each element in \( g \), let \( \ell = (\ell_1, \dots, \ell_d) \in \mathbb{Z}_{p_1} \times \dots \times \mathbb{Z}_{p_d} \) represent the discrete frequency modes in the Fourier space, where \( \ell_j \in \mathbb{Z}_{p_j} \) corresponds to the frequency components in the \( j \)-th dimension, and \( p_j \) is the number of discrete points along the \( j \)-th dimension. The spatial domain is represented by \( y = (y_1, \dots, y_d) \in D \), where \( D \) denotes the computational domain.

Using the Fast Fourier Transform (FFT), the function \( g \) in the spatial domain can be mapped to the frequency domain. The relationship between the Fourier transform \( \hat{\mathcal{G}} \) and its inverse transform \( \hat{\mathcal{G}}^{-1} \) is given by the following equations.

First, the Fourier transform \( \hat{\mathcal{G}} \) is defined as Eq.\ref{eq:fft_forward}:

\begin{equation}
\label{eq:fft_forward}
(\hat{\mathcal{G}} g)_t(\ell) = \sum_{y_1=0}^{p_1-1} \dots \sum_{y_d=0}^{p_d-1} g_t(y_1, \dots, y_d) e^{-2\pi i \sum_{j=1}^d \frac{y_j \ell_j}{p_j}},
\end{equation}

where \( (\hat{\mathcal{G}} g)_t(\ell) \) represents the Fourier transformed data for the \( t \)-th channel of \( g \), and \( \ell \) is the discrete frequency mode in the Fourier space. This equation computes the Fourier transform by summing over all spatial points \( (y_1, \dots, y_d) \), weighted by the Fourier basis function \( e^{-2\pi i \sum_{j=1}^d \frac{y_j \ell_j}{p_j}} \), which relates the spatial coordinates to the frequency modes. The term \( \ell_j \) represents the frequency components in the \( j \)-th dimension, and \( p_j \) is the number of discretized points along that dimension.

Conversely, the inverse Fourier transform \( \hat{\mathcal{G}}^{-1} \) is given by Eq.\ref{eq:fft_inverse}:

\begin{equation}
\label{eq:fft_inverse}
(\hat{\mathcal{G}}^{-1} g)_t(y) = \sum_{\ell_1=0}^{p_1-1} \dots \sum_{\ell_d=0}^{p_d-1} g_t(\ell_1, \dots, \ell_d) e^{2\pi i \sum_{j=1}^d \frac{y_j \ell_j}{p_j}},
\end{equation}

\( (\hat{\mathcal{G}}^{-1} g)_t(y) \) represents the data for the \( t \)-th channel after performing the inverse Fourier transform, and it is computed by summing over all discrete frequency modes \( (\ell_1, \dots, \ell_d) \). The inverse transformation is weighted by the Fourier basis function \( e^{2\pi i \sum_{j=1}^d \frac{y_j \ell_j}{p_j}} \), which maps the frequency components back to the spatial domain.

Eq.\ref{eq:fft_forward} and Eq.\ref{eq:fft_inverse} demonstrate the bidirectional mapping process between the spatial and frequency domains. With the help of the Fast Fourier Transform (FFT), data from the spatial domain is efficiently transformed into the frequency domain, while the inverse Fourier transform brings the data back to the spatial domain. This transformation enables efficient computation of complex spatial operations in the frequency domain, significantly improving computational efficiency in CFD models.





\section*{Appendix B: Test in Montreal city}
\label{sec.quadrantB}

As test cases summarized in Table \ref{tab:cases_summary_combined_appendixMontreal}, to examine the impact of different segmentation strategies on the FNO model’s ability to learn from complex urban wind fields. Fig.B.1 (a) and Fig.B.1 (b) 
compare the segmentation strategies employed for training the FNO model on Montreal's wind field. Panel (a) depicts the whole-field training approach, where the entire wind field is treated as a single, continuous input during training. Panel (b) illustrates the patch-based approach, where the wind field is divided into smaller patches, enabling the FNO model to focus on localized features and complex geometric variations within specific regions.

\begin{figure}[H]
\centering
\begin{picture}(0,0) 
    \put(10,205){\makebox(0,0)[lt]{(a)}}  
\end{picture}
\includegraphics[scale=0.6]{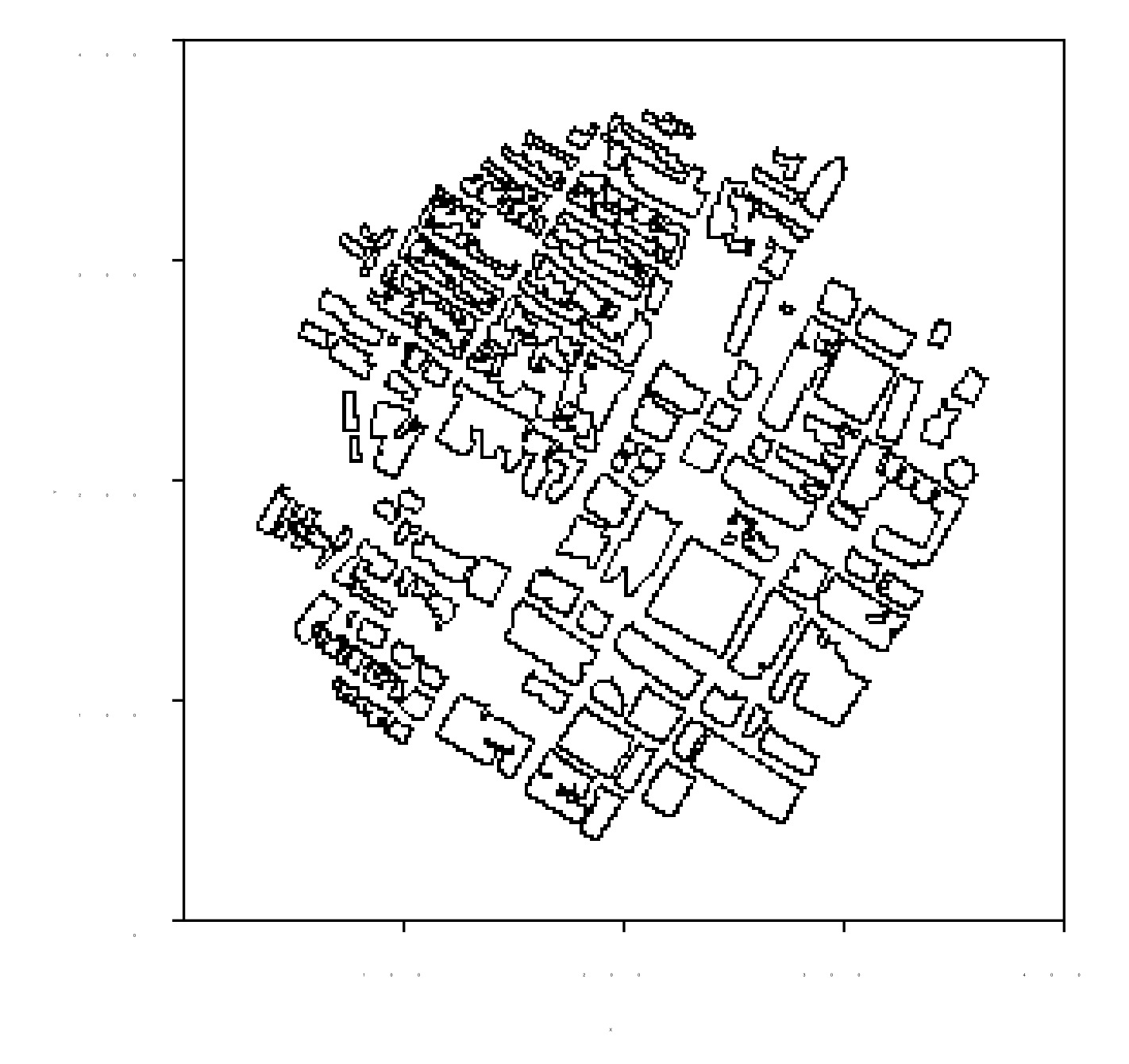}
\begin{picture}(0,0) 
    \put(10,205){\makebox(0,0)[lt]{(b)}}  
\end{picture}
\includegraphics[scale=0.6]{WindEng-elsarticle-V1/figures/montreal2dGridPatchesTraining.jpg}
\caption*{Fig. B.1 Comparison of the Montreal city layout in two scenarios: \textbf{(a)} \textit{W-Mon-T} and \textbf{(b)} \textit{W-Mon-P} for FNO-based generalization tasks.}
\label{Fig.montrealPatchesWholeDivision}
\end{figure}

\begin{table}[H]
\centering
\caption{Summary of cases across different training methods tested on Montreal geometry.}
\fontsize{1.0pt}{1.2pt}\selectfont

\setlength{\heavyrulewidth}{0.03mm}   
\setlength{\lightrulewidth}{0.015mm}  
\setlength{\aboverulesep}{0.15pt}     
\setlength{\belowrulesep}{0.15pt}     
\setlength{\arrayrulewidth}{0.1mm}   

\resizebox{\textwidth}{!}{
\begin{tabular}{c@{\extracolsep{5pt}}c@{\extracolsep{5pt}}c}
\toprule
\textbf{Training Methods} & \textbf{Cases}  & \textbf{Instruction} \\

\midrule
Train Dataset
                          & \textit{W-Nii-T-CFD} & Train in total domain without SDF\\
                          & \textit{W-Nii-T-SDF-CFD} & Train in total domain with SDF\\
                          & \textit{W-Nii-P-CFD} & Train in patches without SDF\\
                          & \textit{W-Nii-P-SDF-CFD} & Train in patches with SDF\\
\midrule
Test Cases
                          & \textit{W-Mon-T} & Test in total domain without SDF \\
                          & \textit{W-Mon-T-SDF}  & Test in total domain with SDF \\
                          & \textit{W-Mon-P} & Test in patches without SDF\\
                          & \textit{W-Mon-P-SDF}  & Test in patches with SDF\\

\bottomrule

\end{tabular}
}

\vspace{10pt}
\parbox{\textwidth}{\footnotesize
\textbf{*} \textbf{\textit{W}} in the first column stands for West wind direction, \textbf{\textit{N}} for North wind direction, \textbf{\textit{Nii}} for Niigata, \textbf{\textit{Mon}} for Montreal, \textbf{\textit{SDF}} for Signed distance function data, \textbf{\textit{CFD}} for Computational fluid dynamics data, \textbf{\textit{P}} for Patches blocks, \textbf{\textit{T}} for total domain, \textbf{\textit{R}} for Rotated 90 degrees counterclockwise, \textbf{\textit{VF}} for Vertically flipped. 
}

\label{tab:cases_summary_combined_appendixMontreal}
\end{table}

\begin{figure}
\centering
    \includegraphics[scale=0.1]{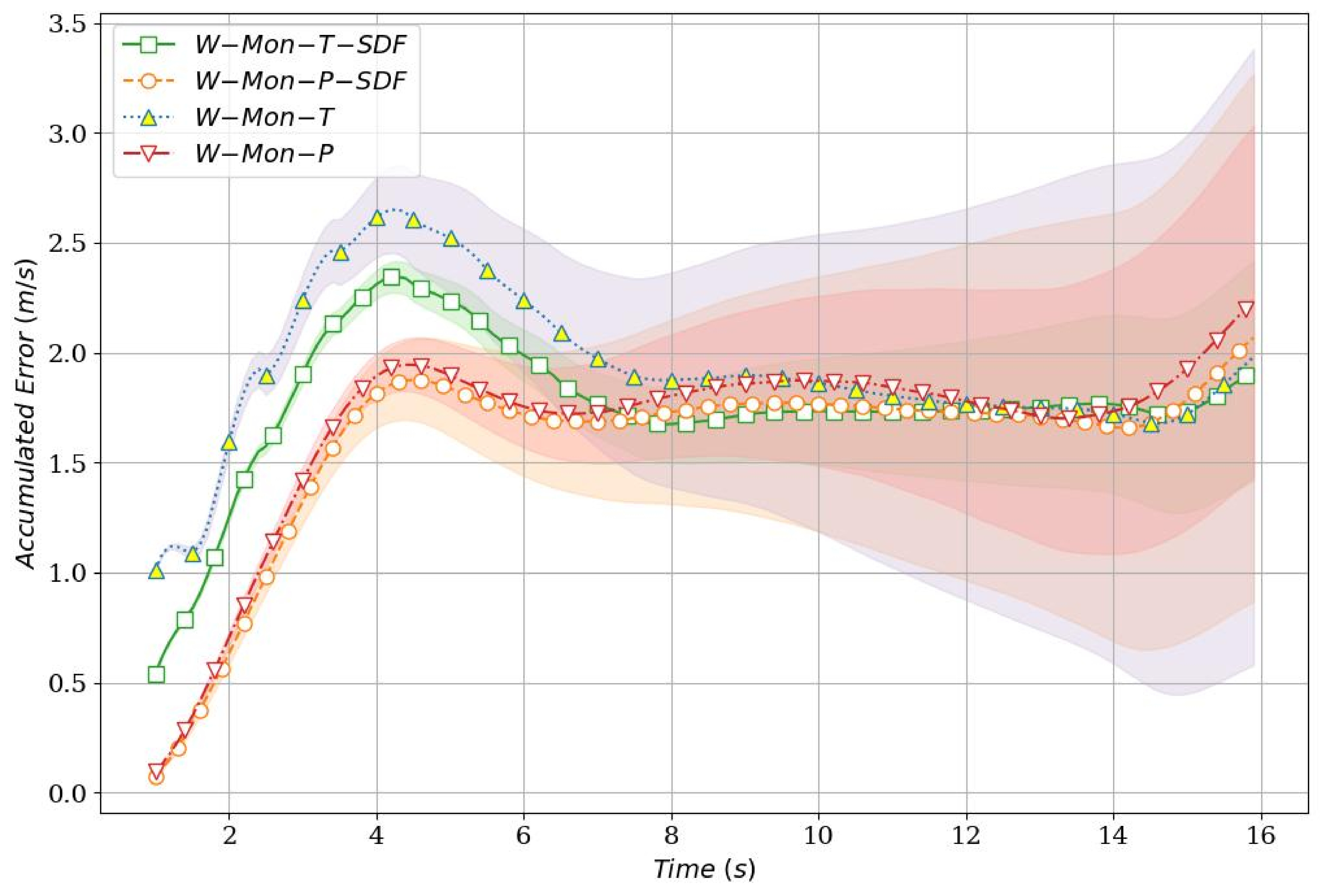}
\caption*{Fig C.1 Comparison of the accumulated average absolute error in Montreal over time on different training methods.}
\label{Fig.MontrealPatchesWholeLineGraph}
\end{figure}

Fig.C.1 shows a comparison of the accumulated average absolute error over time for wind field predictions in Montreal using different training methods: \textit{W-Mon-T-SDF}, \textit{W-Mon-P-SDF}, \textit{W-Mon-T}, and \textit{W-Mon-P}. The results indicate that patch-based training methods, particularly \textit{W-Mon-P-SDF}, exhibit lower and more stable error accumulation throughout the time steps, with a slower rate of error growth, especially in longer-term predictions. In contrast, whole-field training methods (\textit{W-Mon-T} and \textit{W-Mon-T-SDF}) show higher initial errors and faster accumulation of errors. Although patch-based methods theoretically provide the best prediction performance, the complex urban geometry of Montreal leads to larger energy capture errors for this approach. This highlights the significant impact of structural differences on model performance, emphasizing the importance of geometric consistency for accurate predictions, especially when there is a large disparity between training data and the test city's layout.

\begin{table}
\centering
\caption{Comparison of Accumulated Radial Energy Spectrum Differences for Different Training Methods and SDF Impact tested in Montreal.}

\fontsize{1.0pt}{1.2pt}\selectfont

\setlength{\heavyrulewidth}{0.03mm}   
\setlength{\lightrulewidth}{0.015mm}  
\setlength{\aboverulesep}{0.15pt}     
\setlength{\belowrulesep}{0.15pt}     
\setlength{\arrayrulewidth}{0.1mm}   

\resizebox{\textwidth}{!}{
\begin{tabular}{c@{\extracolsep{5pt}}c@{\extracolsep{5pt}}c@{\extracolsep{5pt}}c@{\extracolsep{5pt}}c}
\toprule
\multicolumn{5}{c}{\textbf{Accumulated Error at Selected Time Steps (m/s)}} \\
\midrule
Time Step & \textit{W-Mon-P-SDF} & \textit{W-Mon-P} & \textit{W-Mon-T} & \textit{W-Mon-T-SDF}\\
\midrule
5   & 0.26 & 0.29 & 1.08 & 0.79 \\
10  & 0.57 & 0.63 & 1.27 & 1.16 \\
15  & 0.91 & 1.01 & 1.88 & 1.55 \\
20  & 1.26 & 1.35 & 2.07 & 1.83 \\
\midrule
\multicolumn{5}{c}{\textbf{Radial Energy Spectrum Absolute Differences at Selected Wave Numbers}} \\
\midrule
Wave Number & \textit{W-Mon-P-SDF} & \textit{W-Mon-P} & \textit{W-Mon-T} & \textit{W-Mon-T-SDF} \\
\midrule
32   & 229.27 & 283.57 & 309.22 & 253.74 \\
64   & 67.53  & 93.06  & 90.61  & 79.27  \\
96   & 23.35  & 41.93  & 35.17  & 35.29  \\
128  & 12.01  & 26.09  & 21.29 & 20.46  \\
\bottomrule
\end{tabular}
}

\label{tab:combined_accumulated_error_spectrum_montreal}
\end{table}

For further assessing how training approaches impact both temporal error accumulation and spectral energy fidelity in complex urban wind dynamics, Table \ref{tab:combined_accumulated_error_spectrum_montreal} provides a comparative analysis of accumulated errors and radial energy spectrum absolute differences between \textit{W-Nii-T-CFD}, \textit{W-Nii-T-SDF-CFD}, \textit{W-Nii-P-CFD} and \textit{W-Nii-T-CFD} cases for the Montreal wind field. Accumulated errors (m/s) at the selected time (0.5 s, 1 s, 1.5 s, and 2 s) are shown to be consistently lower for the \textit{W-Mon-P-SDF} case than the other three cases. The patch-based training method, by dividing the wind field into smaller local regions and incorporating SDF information, is able to capture the dynamic variations of the localized wind field more effectively, which is crucial for accurate wind field prediction. While this approach aids the FNO model in identifying localized features and effectively managing error accumulation during training, its performance in Montreal is significantly lower than in Niigata. This highlights the importance of geometric similarity between the training and testing environments for prediction accuracy. Although the patch-based method shows promising results for localized wind field predictions in Montreal, the substantial difference in urban geometry between Montreal and Niigata hinders the FNO model's ability to effectively transfer the spatial features learned during training, leading to higher error accumulation. In contrast, Niigata's wind field shares a high degree of geometric similarity with the training data, allowing the patch-based method to capture local variations more effectively and maintain lower errors in prediction. Therefore, these results further emphasize the importance of geometric similarity between the training and target cities, suggesting that wind field predictions are more accurate when applied to cities with similar geometric structures.
Absolute differences in the radial energy spectrum are compared at wave numbers 32, 64, 96, and 128. At the lower wave number of 32, an absolute difference of 229.27 is noted for the \textit{W-Mon-P-SDF} case, while the \textit{W-Mon-T} case shows a larger difference of 309.22. As wave numbers increase, absolute differences decrease for all cases; however, higher errors are consistently observed for \textit{W-Mon-T}. In Montreal, the complexity of the urban geometry significantly differs from that of Niigata, making it difficult for the FNO model to effectively transfer the localized spatial features learned during training to the new city layout. In contrast, when the training data and the test city share similar geometric structures (as in the case of Niigata), the FNO model is better able to adapt to the localized structural features, leading to more accurate predictions. Therefore, this phenomenon underscores the importance of geometric consistency in wind field simulations.

\section*{Appendix C. Radial Energy Spectrum}

We describe the procedure used to compute the radial energy spectrums in sections above. Unlike the standard radially averaged energy spectrum, our approach ensures the preservation of total energy.

\begin{enumerate}
 \item \textbf{Application of 2-Dimensional Fourier Transform}: Adopt the Fast Fourier Transform (FFT) to transform the input 2D matrix from the spatial domain to the frequency domain. \item \textbf{Determine wave numbers}: For each element in the transformed matrix, calculate its corresponding wave number, which is defined as the distance from the matrix’s center to that element. \item \textbf{Divide wave numbers into bins}: Group the computed wave numbers into discrete intervals (bins). For a 64 $\times$ 64 matrix, we create 32 bins, with each bin covering a wave number range of 1. \item \textbf{Computing energy per bin}: Sum the squared magnitudes of the Fourier coefficients that fall into that bin’s wave number range within each bin. This sum represents the total energy associated with that frequency range. 
\end{enumerate}

The radial energy spectrum redistributes the 2D spectrum's energy radially, resulting in a clearer visualization.

To calculate the energy for each bin, we start by defining the Fourier transform of the input matrix as \( \hat{f}(k_x, k_y) \), where \( k_x \) and \( k_y \) are the wave numbers in the frequency domain. The energy at each frequency point is given by the squared magnitude of the Fourier coefficient:

\[
E(k_x, k_y) = |\hat{f}(k_x, k_y)|^2
\]

Next, we compute the radial distance \( r \) of each frequency point from the center of the frequency space:

\[
r = \sqrt{k_x^2 + k_y^2}
\]

The Fourier coefficients are then binned according to their radial distance \( r \). For each bin \( i \), the total energy \( E_i \) is calculated as the sum of squared magnitudes of the Fourier coefficients within that bin:

\[
E_i = \sum_{(k_x, k_y) \in B_i} |\hat{f}(k_x, k_y)|^2
\]

where \( B_i \) denotes the set of wave numbers \( (k_x, k_y) \) that fall into the \( i \)-th bin, and \( E_i \) represents the total energy in that bin.

This method redistributes the 2D spectral energy radially, making it easier to visualize the energy distribution across different wavenumber ranges.

\section*{Appendix D. SDF calculation}

The SDF is a mathematical representation that encodes the minimum distance from any given point in a domain to the nearest boundary of a defined geometry, with a sign indicating whether the point lies inside or outside the boundary. It is widely used in CFD and machine learning for spatial representation, especially in cases involving complex urban geometries. To compute the SDF, the following procedure is typically followed :

Binary Representation of the Geometry: The input geometry, such as a wind field or urban layout, is first discretized into a binary matrix, where 1 indicates the presence of obstacles (e.g., buildings), and 0 represents free-flow regions.
Distance Calculation: The Euclidean distance is computed for each point in the domain:
For points in the free-flow region (0), the distance is calculated to the nearest obstacle.
For points inside the obstacles (1), the distance is calculated to the nearest boundary of the obstacle.
Sign Assignment: The computed distances are assigned a positive or negative sign:
A negative sign is applied to points within obstacles.
A positive sign is assigned to points outside obstacles.
Combining Results: The final SDF is generated by merging the signed distances from both the free-flow and obstacle regions, creating a continuous field that seamlessly represents the geometry.
This approach ensures that the SDF captures the spatial relationships and boundary conditions of the geometry effectively, enabling its use in deep learning models for tasks like urban wind field simulation. By incorporating SDF data, models can leverage additional geometric information, improving prediction accuracy and stability.

\begin{enumerate}
    \item \textbf{Binary Representation of the Geometry:} The input geometry, such as a wind field or urban layout, is first discretized into a binary matrix, where 1 indicates the presence of obstacles (e.g., buildings), and 0 represents free-flow regions.
    \item \textbf{Distance Calculation:} The Euclidean distance is computed for each point in the domain. For each point \( p = (x_1, x_2, \dots, x_d) \in D \), the distance \( d(p) \) to the nearest boundary is computed as:
    \[
    d(p) = \min_{q \in \partial \Omega} \| p - q \|
    \]
    where \( \partial \Omega \) represents the boundary of the domain (obstacle or free-flow region), and \( \| p - q \| \) is the Euclidean distance between the point \( p \) and the closest boundary point \( q \).
    \item \textbf{Sign Assignment:} The computed distances are assigned a positive or negative sign:
    \begin{itemize}
        \item A negative sign is applied to points inside obstacles.
        \item A positive sign is assigned to points outside obstacles.
    \end{itemize}
    Mathematically, the SDF can be written as:
    \[
    SDF(p) = 
    \begin{cases} 
    -d(p) & \text{if } p \in \text{Obstacle Region} \\
    d(p) & \text{if } p \in \text{Free-flow Region}
    \end{cases}
    \]
    where \( SDF(p) \) is the signed distance at point \( p \).
    \item \textbf{Combining Results:} The final SDF is generated by merging the signed distances from both the free-flow and obstacle regions, creating a continuous field that seamlessly represents the geometry. This ensures the signed distance captures both the shape and spatial relationship of the boundary effectively.
\end{enumerate}

\section*{Appendix E. SSIM Calculation}

The Structural Similarity Index (SSIM)\cite{bakurov2022structural} is a metric proposed to evaluate the similarity between two pictures by considering luminance, contrast, and structural information. Its formula is given by:

\[
\text{SSIM}(x, y) = \frac{(2\mu_x\mu_y + C_1)(2\sigma_{xy} + C_2)}{(\mu_x^2 + \mu_y^2 + C_1)(\sigma_x^2 + \sigma_y^2 + C_2)}
\]

where \( \sigma_x^2 \) and \( \sigma_y^2 \) are their variances, \( \mu_x \) and \( \mu_y \) represent the mean intensities of images \( x \) and \( y \), and \( \sigma_{xy} \) is the covariance. The constants \( C_1 \) and \( C_2 \) are included to avoid instability when the denominator is close to zero. SSIM values range from -1 to 1, where a value of 1 indicates perfect similarity.

\label{ref}
\bibliography{WindEng-elsarticle-V1/manuscript}

\bibliographystyle{model3-num-names}

\end{document}